\documentclass{ieeeaccess}
\usepackage{cite}
\usepackage{amsmath,amssymb,amsfonts}
\usepackage{graphicx}
\usepackage{textcomp}
\usepackage{caption}
\usepackage{framed,multirow}
\usepackage{subfigure}
\usepackage{algorithm}
\usepackage{algpseudocode}
\numberwithin{equation}{section}
\usepackage{latexsym}
\usepackage{booktabs}
\usepackage{array}
\usepackage{comment}
\usepackage{multirow}

\usepackage{amsthm}
\theoremstyle{definition}

\newtheorem{mydeff}{Constraint}
\newtheorem{mydefff}{Example}
\usepackage{color}
\usepackage{soul}
\soulregister\cite7
\soulregister\ref{7}
\usepackage{enumitem}
\def\BibTeX{{\rm B\kern-.05em{\sc i\kern-.025em b}\kern-.08em
		T\kern-.1667em\lower.7ex\hbox{E}\kern-.125emX}}
\begin{document}
	\history{Date of publication xxxx 00, 0000, date of current version xxxx 00, 0000.}
	\doi{00.0000/ACCESS.2020.DOI}
	
	\title{Reference-Based Sequence Classification}
	
	\author{\uppercase{Zengyou He}\authorrefmark{1},
            \uppercase{Guangyao Xu}\authorrefmark{1},
            \uppercase{Chaohua Sheng}\authorrefmark{1},
            \uppercase{Bo Xu}\authorrefmark{1},
		\uppercase{Quan Zou}\authorrefmark{2}.}
	
	\address[1]{School of Software, Dalian University of Technology, Tuqiang Road, Dalian, China}
	\address[2]{Institute of Fundamental and Frontier Sciences, University of Electronic Science and Technology, Chengdu, China}
	
	\corresp{Corresponding author: Zengyou He (e-mail: zyhe@dlut.edu.cn).}
	\tfootnote{This work was partially supported by the Natural Science Foundation of China under Grant Nos. 61972066 and 61572094, and
the Fundamental Research Funds for the Central Universities (No. DUT20YG106).}
	
	\begin{abstract}
		Sequence classification is an important data mining task in many real-world applications. Over the past few decades, many sequence classification methods have been proposed from different aspects. In particular, the pattern-based method is one of the most important and widely studied sequence classification methods in the literature. In this paper, we present a reference-based sequence classification framework, which can unify existing pattern-based sequence classification methods under the same umbrella. More importantly, this framework can be used as a general platform for developing new sequence classification algorithms. By utilizing this framework as a tool, we propose new sequence classification algorithms that are quite different from existing solutions. Experimental results show that new methods developed under the proposed framework are capable of achieving comparable classification accuracy to those state-of-the-art sequence classification algorithms.
	\end{abstract}
	
	
	\titlepgskip=-15pt
	
	\maketitle
	%\linenumbers
	
	\section{Introduction}
\label{sec:1}
\PARstart{I}{n} many practical applications, we have to conduct data analysis on data sets that are composed of discrete
sequences. Each sequence is an ordered list of elements. For instance, such a sequence can be a protein sequence, where each
element corresponds to an amino acid. Due to the existence of a large number of discrete sequences in a wide range of applications, sequential data analysis has become an important issue in machine learning and data mining. Compared to non-sequential data mining, sequential data analysis is confronted with new challenges because of the ordering relationship between different elements in the sequences. Similar to the analysis of non-sequential data, there are different sequential data mining problems such as clustering, classification and pattern discovery. In this paper, we focus on the sequence classification problem.

The task of classification is to determine which predefined target class one unknown object should be assigned to~\cite{han2011data}. As a specific case of the general classification problem, sequence classification is to assign class labels to new sequences based on the classifier constructed in the training phase. In many real-world applications, we can formulate the data analysis task as a sequence classification problem. For instance, the essential task in numerous bioinformatics applications is to classify biological sequences into existing categories~\cite{deshpande2002evaluation}.

To tackle the sequence classification problem, many effective methods have been proposed from different aspects. Roughly, existing sequence classification methods can be divided into three categories~\cite{Xing2010A}: feature-based methods, distance-based methods and model-based methods. Feature-based methods first transform sequences into feature vectors and then apply existing vectorial data classification methods. Distance-based methods apply classifiers such as KNN ($k$ Nearest Neighbors) to solve the sequence classification problem, in which the key issue is to specify a proper distance function to measure the distance between two sequences~\cite{Xing2010A}. Model-based methods generally assume that sequences from different classes are generated from different probability distributions, in which the key issue is to estimate the model parameters from the set of training sequences.

In this paper, we focus on the feature-based method since it has several advantages. First of all, various effective classifiers have been developed for vectorial data classification~\cite{Cernadas2014Do}. After transforming sequences into feature vectors, we can choose any one of these existing classification methods to fulfill the sequence classification task. Second, in some popular feature-based methods such as pattern-based methods, each feature has a good interpretability. Last but not least, the extraction of features from sequences has been extensively studied across different fields, making it feasible to generate sequence features in an effective manner.

The $k$-mer (in bioinformatics) or $k$-gram (in natural language processing) is a substring that is composed of
$k$ consecutive elements, which is probably the most widely used feature in feature-based sequence classification. Such a $k$-mer based feature construction method is further generalized by the pattern-based method, in which a feature is a sequential pattern (a subsequence) that satisfies some constraints (e.g. frequent pattern, discriminative pattern). Over the past few decades, a large number of pattern-based methods have been presented in the context of sequence classification~\cite{zhou2016pattern,exarchos2008two,lo2009classification,
she2003frequent,hopf2010mining,haleem2014novel,deng2010occurrence,deng2009contrasting,he2019significance,
fradkin2015mining,yahyaoui2016feature,lee2015multi,egho2017user,ntagiou2017protein,tsai2015pso,
batal2013temporal,tsai2013time,exarchos2009optimized,tseng2005cbs,tseng2009effective,
syed2009learning,lesh1999mining,rani2008rbnbc,holat2014sequence,febrer2016spac,he2018mining}.

In this paper, we present a reference-based sequence classification framework, which can be considered as a non-trivial generalization of the pattern-based methods. This framework has several key steps: candidate set construction, reference point selection and feature value construction.
In the first step, a set of sequences that serve as the candidate reference points are constructed. Then, some sequences from the candidate set are selected as the reference points according to certain criteria. The number of features in the transformed vectorial data will equal the number of selected reference points. In other words, each reference point will correspond to a transformed feature. Finally, a similarity function is
used to calculate the similarity between each sequence in the data and every reference point. The similarity to each reference point will be used as the corresponding feature value.

The reference-based sequence classification framework is quite general and flexible since the selection of both reference points and similarity functions is arbitrary. Existing feature-based methods can be regarded as a special variant under our framework by (1) using (frequent or discriminative) sequential patterns (subsequences) as reference points and (2) utilizing a boolean function (output 1 if the reference point is contained in a given sequence and output 0 otherwise) as the similarity function. Besides unifying existing pattern-based methods under the same umbrella, the reference-based sequence classification framework can be used as a general platform for developing new feature-based sequence classification methods. To justify this point, we develop a new feature-based method in which a subset of training sequences are used as the reference points and the Jaccard coefficient is used as the similarity function. In particular, we present two instance selection methods to select a good set of reference points.

To demonstrate the feasibility and advantages of this new framework, we conduct a series of comprehensive performance studies on real sequential data sets. In the experiments, we compare several variants under our framework with some existing sequence classification methods in terms of classification accuracy. Experimental results show that new methods developed under the proposed framework are capable of achieving better classification accuracy than traditional sequence classification methods. This indicates that such a reference-based sequence classification framework is promising from a practical point of view.

The main contributions of this paper can be summarized as follows:
\begin{itemize}
\item We present a general reference-based framework for feature-based sequence classification. It offers a unified view for understanding and explaining many existing feature-based sequence classification methods in which different types of sequential patterns are used as features.
\item The reference-based framework can be used as a general platform for developing new feature-based sequence classification algorithms. To verify this point, we design new feature-based sequence classification algorithms under this framework and demonstrate its advantages through extensive experimental results on real sequential data sets.
\end{itemize}

The rest of the paper is structured as follows. Section \ref{sec:2} gives a discussion on the related work. In Section \ref{sec:3}, we introduce the reference-based sequence classification framework in detail. In Section \ref{sec:4}, we show that many existing feature-based sequence classification algorithms can be reformulated within the reference-based framework. In Section \ref{sec:5}, we present new feature-based sequence classification algorithms under this framework, which are effective and quite different from available solutions. We experimentally evaluate the proposed reference-based framework through a series of experiments on real-life data sets in Section \ref{sec:6}. Finally, we summarise our research and give a discussion on the future work in Section \ref{sec:7}.

\section{Related Work}
\label{sec:2}
In this section, we discuss previous research efforts that are closely related to our method. In Section \ref{sec:2.1}, we provide a categorization on existing feature-based sequence classification methods. In Section \ref{sec:2.2}, we discuss several instance-based feature generation methods in the literature of time series classification. In Section \ref{sec:2.3}, we present a concise discussion on reference-based sequence clustering algorithms. In Section \ref{sec:2.4}, we provide a short summary on dimension reduction and embedding methods based on landmark points.

\subsection{Feature-Based Methods}
\label{sec:2.1}

\subsubsection{Explicit Subsequence Representation without Selection}
\label{sec:2.1.1}
The naive approach in dealing with discrete sequences is to treat each element as a feature. However, the order information between different elements
will be lost and the sequential nature cannot be captured in the classification. Short sequence segments
of $k$ consecutive elements called $k$-grams can be used as features to solve this problem. Given a set of
$k$-grams, a sequence can be represented as a vector of the presence or absence of the $k$-grams or
the frequencies of the $k$-grams. In this feature representation method, all $k$-grams (for a specified $k$ value) are explicitly used as the features without feature selection.

\subsubsection{Explicit Subsequence Representation with Selection (Classifier-Independent)}
\label{sec:2.1.2}
Lesh et al.~\cite{lesh1999mining} present a pattern-based classification method in
which a sequential pattern is chosen as a feature. The selected pattern should satisfy the following criteria: (1) be frequent, (2) be
distinctive of at least one class and (3) not redundant. Towards this direction, many pattern-based classification methods have been subsequently proposed, in which different constraints are imposed on the patterns that should be selected as features~\cite{zhou2016pattern,exarchos2008two,lo2009classification,
she2003frequent,hopf2010mining,haleem2014novel,deng2010occurrence,deng2009contrasting,he2019significance,
fradkin2015mining,yahyaoui2016feature,lee2015multi,egho2017user,ntagiou2017protein,tsai2015pso,
batal2013temporal,tsai2013time,exarchos2009optimized,tseng2005cbs,tseng2009effective,
syed2009learning,rani2008rbnbc,holat2014sequence,febrer2016spac,he2018mining}. Note that any classifier designed for vectorial data can be applied to the transformed data generated from such pattern-based methods. In other words, such feature generation methods are classifier-independent.

\subsubsection{Explicit Subsequence Representation with Selection (Classifier-Dependent)}
\label{sec:2.1.3}
The above pattern-based methods are universal and classifier-independent. However, some patterns that are critical to the classifier may be
filtered out during the selection process. Thus, several methods which can select pattern features from the entire pattern space for a specific classifier have been proposed~\cite{ifrim2008fast,ifrim2011bounded,okanohara2009text}.

In~\cite{ifrim2008fast}, a coordinate-wise gradient ascent technique is presented for learning the logistic regression function in the space of all $k$-grams. The method exploits the inherent structure of the $k$-gram feature space to automatically provide a compact set of highly discriminative $k$-gram features. In~\cite{ifrim2011bounded}, a framework is presented in which linear classifiers such as logistic regression
and support vector machine can work directly in the explicit high-dimensional space of all subsequences. The key idea is a gradient-bounded coordinate-descent strategy to quickly retrieve features without explicitly enumerating all potential subsequences. In~\cite{okanohara2009text}, a novel document classification method using all substrings as features is proposed, in which $L_1$ regularization is applied to a multi-class logistic regression model to fulfill the feature selection task automatically and efficiently.

\subsubsection{Implicit Subsequence Representation}
\label{sec:2.1.4}
In contrast to explicit subsequence representation, kernel-based methods employ an implicit subsequence representation strategy. A kernel function is the key ingredient for learning with support vector machines (SVMs) and it implicitly defines a high-dimensional feature space. Some kernel functions $K(x,y)$ have been presented for measuring the
similarity between two sequences $x$ and $y$ (e.g.~\cite{sonnenburg2005learning}).

There are a variety of string kernels which are widely used for sequence classification (e.g.~\cite{leslie2001spectrum,lodhi2002text,eskin2003mismatch,leslie2004fast}).
A sequence is transformed into a feature space and the kernel function is the inner product of two transformed feature vectors.

Leslie et al.~\cite{leslie2001spectrum} propose a $k$-spectrum kernel for protein
classification. Given a number $k \geq 1$, the $k$-spectrum of an input sequence
is the set of all its $k$-length (contiguous) subsequences.

Lodhi et al.~\cite{lodhi2002text} present a string kernel based on gapped $k$-length subsequences for text classification. The subsequences are weighted
by an exponentially decaying factor of their full length in the text.

In~\cite{eskin2003mismatch}, a mismatch string kernel is proposed, in which a certain number of mismatches are allowed in counting the occurrence of a subsequence. Several string kernels related to the mismatch kernel are presented in~\cite{leslie2004fast}: restricted gappy kernels, substitution kernels and wildcard kernels.
			
\subsubsection{Sequence Embedding}
\label{sec:2.1.5}
All the methods mentioned above use subsequences as features. Alternatively, the sequence embedding method generates a vector representation in which each feature does not have a clear interpretation. Most existing approaches for sequence embedding are proposed for texts in natural language processing, where word and document embeddings are used as an efficient way to encode the text (e.g.~\cite{mikolov2013distributed,le2014distributed}). The basic assumption in these methods is that words that appear in similar contexts have similar meanings.

The word2vec model~\cite{mikolov2013distributed} uses a two-layer neural network to learn a vector representation for each word. The sequence (text) embedding vector can be further generated by combining the feature vectors for words. The doc2vec model~\cite{le2014distributed} extends word2vec by directly learning feature vectors for entire sentences, paragraphs, or documents.

Nguyen et al.~\cite{nguyen2018sqn2vec} propose an unsupervised method (named Sqn2Vec) for learning sequence embedding by predicting its belonging singleton symbols and sequential patterns (SPs). The main objective of Sqn2Vec is to address the limitations of two existing approaches: pattern-based methods often produce sparse and high-dimensional feature vectors while sequence embedding methods in natural language processing may fail on data sets with a small vocabulary.

\subsubsection{Summary of Feature-Based Methods}
\label{sec:2.1.6}
Roughly, existing feature-based sequence classification methods can be divided into the above five categories. Each of these methods has its pros and cons, which we will discuss briefly next.

First, using $k$-grams as features without feature selection is simple and effective in practice. However, the feature length $k$ cannot be large and many redundant features may be included.

Second, in the pattern-based method, the length of a feature is not restricted as long as the feature satisfies given constraints and redundant features can be filtered out in some formulations. However, it is a non-trivial task to efficiently mine patterns that can satisfy the constraints.

Third, sequence classification methods based on adaptive feature selection can automatically select features from the set of all subsequences. The basic idea is to integrate the feature selection and classifier construction into the same procedure. Hence, these methods are classifier-dependent in the sense that each algorithm is only applicable to a specific classifier.

Fourth, kernel-based methods can implicitly map the sequence into a high-dimensional feature space without explicit feature extraction. The major challenge is how to choose a proper string kernel function and how to handle large data sets efficiently.

Finally, sequence embedding methods generate a new vector representation for each sequence that may achieve better classification accuracy. Unfortunately, the semantic interpretation of each feature becomes a difficult issue.

\subsection{Instance-Based Feature Generation Methods}
\label{sec:2.2}
There are several instance-based feature generation methods for time series classification which are closely related to our method (e.g.~\cite{iosifidis2013multidimensional,kate2016using}).

Iosifidis et al.~\cite{iosifidis2013multidimensional} propose a time series classification method based on a novel vector representation. The vector representation for each time series is generated by calculating its similarities from a subset of training instances. To find a good subset of representative instances, a clustering procedure is further presented. In~\cite{kate2016using}, each time series is represented as a feature vector, where the feature value is its dynamic time warping similarity from one of the training instances. Note that all training instances are used for feature generation.

\subsection{Reference-Based Sequence Clustering}
\label{sec:2.3}
In the literature of sequence clustering, the idea of using reference/landmark points to accelerate the cluster analysis process has been widely studied (e.g.~\cite{blackshields2008fast,voevodski2012active}). In this type of sequence clustering algorithm, a reference point selection method is first employed to obtain a small set of landmark points and then the clustering process is conducted based on the similarities between input sequences and selected reference points. Here, we would like to highlight the following differences between our method and existing research efforts in this field: (1) The objective is different. We focus on the classification issue while these methods aim at the cluster analysis problem. Besides, their main concern is to improve the running efficiency of the sequence clustering procedure; (2) The method is different. We present two reference point selection methods: one unsupervised method and one supervised method (see Section \ref{sec:5} for the details). In existing reference-based sequence clustering methods, only the unsupervised reference point selection method is applicable since no class label information is available.

\subsection{Reference-Based Dimension Reduction}
\label{sec:2.4}
A number of research papers have presented the idea of using the distances to a set of reference points to fulfill the dimension reduction task (e.g.~\cite{Faloutsos1995FastMap,hjaltason2003properties}). Our method shares some similarities with these methods since the final objective is the same. However, most of these methods are not developed for the task of sequence classification. As a result, our method is quite different from these methods for both the reference point selection and the similarity computation.

\begin{figure}[htbp]%%图
	\centering  %插入的图片居中表示
	\includegraphics[width=0.48\textwidth]{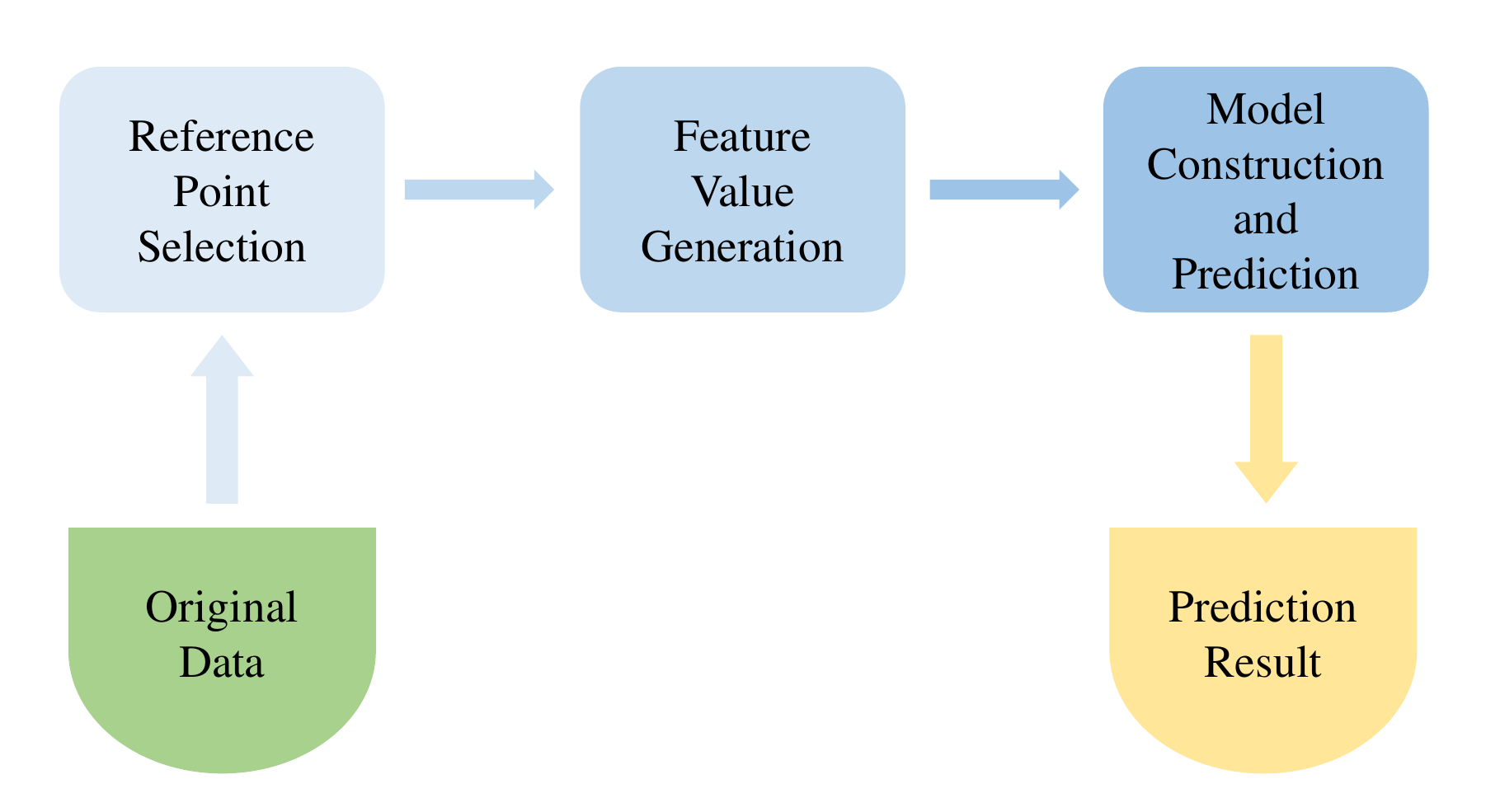}  %插入的图，包括JPG,PNG,PDF,EPS 等，放在源文件目录下
	\caption{The entire workflow of reference-based sequence classification framework}  % 图片的名称
	\label{fig:mcmthesis-logo1}   %标签，用作引用
\end{figure}

\section{Reference-Based Sequence Classification Framework}
\label{sec:3}
Let $I=\left\{i_1,i_2,...,i_m\right\}$ be a finite set of $m$ distinct items, which is generally called the alphabet in the literature. A sequence $s$ over $I$ is an ordered list $s = \left \langle s_1,s_2,...,s_l \right \rangle$, where $s_i \in I$ and $l$ is the length of the sequence $s$. A sequence $t = \left \langle t_1,t_2,...,t_r \right \rangle$ is said to be a subsequence of $s$ if there exist integers $1 \leq i_1 < i_2 <...< i_r \leq l$ such that $t_1 = s_{i_1},t_2 = s_{i_2}, ... ,t_r = s_{i_r}$, denoted as $t \subseteq s$ (if $t \neq s$, written as $t \subset s$). We use $maxsize$ to denote the allowed maximum length of subsequences.

Let $C=\left\{c_1,c_2,...,c_j\right\}$ be a finite set of $j$ distinct classes. A labeled sequential data set $D$ over $I$ is a set of instances and each instance $d$ is denoted by $(s,c_k)$, where $s$ is a sequence and $c_k \in C$ is a class label, $|D|$ is the number of sequences in $D$. The set $D_{c_i} \subseteq D$ contains all sequences that have the same class label $c_i$ (i.e., $D= \cup^j_{i=1} D_{c_i}$). $D_{c_i}(t)$ is the set of sequences in $D_{c_i}$ that contain $t$, where $t$ is a given sequence. Sequences in $D$ ($D_{c_i}$) is divided into a training set $TrainD$ ($TrainD_{c_i}$) and a testing set $TestD$ ($TestD_{c_i}$). The set of all subsequences of $TrainD$ is denoted by $SubTrainD=\left\{t|t\subseteq s, s \in TrainD \right\}$.

As shown in Fig. \ref{fig:mcmthesis-logo1}, we present a reference-based sequence classification framework. It is composed of three major phases: reference point selection, feature value generation, model construction and prediction. In the following, we will elaborate on each step in detail.

\begin{figure*}[htbp]%%图
	\centering  %插入的图片居中表示
	\includegraphics[width=0.85\textwidth]{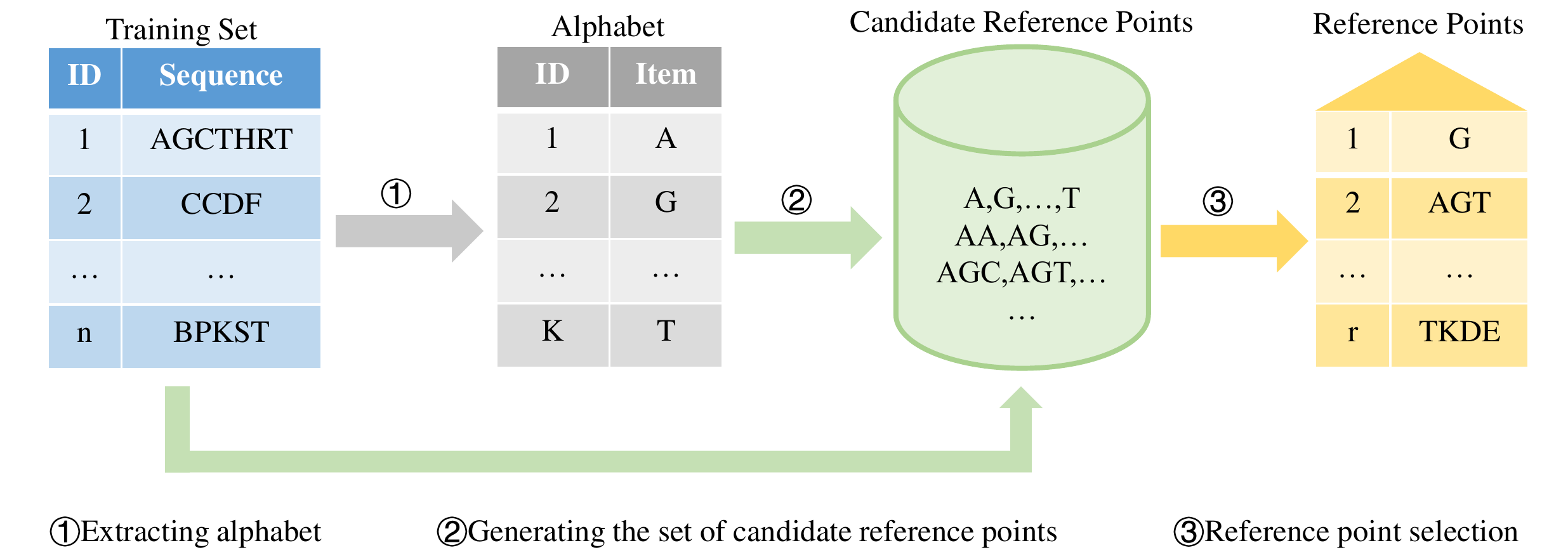}  %插入的图，包括JPG,PNG,PDF,EPS 等，放在源文件目录下
	\caption{The process of reference point selection}  %图片的名称
	\label{fig:mcmthesis-logo2}   %标签，用作引用
\end{figure*}

\subsection{Reference Point Selection}
\label{sec:3.1}
In the first stage of the presented framework, a reference point selection procedure is performed to generate a set of pivot sequences. As shown in Fig. \ref{fig:mcmthesis-logo2}, this procedure can be further divided into three steps: alphabet extraction, candidate set generation and pivot sequence selection.

In the first step, we scan the training set $TrainD$ to extract the alphabet $I$ that is composed of distinct items. Note that there can be some items that only appear in the testing set $TestD$. In the forthcoming paragraphs, we will see that this extreme case does not affect our subsequent steps.

In the second step, we generate the set of candidate reference sequences $CR$ from the alphabet $I$. Note that any sequence over $I$ can be a member of $CR$. In other words, $CR$ can be an infinite set. In practice, some constraints will be imposed on the potential member in $CR$. For instance, those pattern-based methods only consider subsequences of $TrainD$ as members of $CR$ under our framework, which will be further discussed in Section \ref{sec:4}. Furthermore, the use of different construction methods for building the candidate set $CR$ will lead to the generation of many new feature-based sequence classification methods.

In the third step, we select a subset of sequences $R$ from $CR$ as the landmark sequences for generating features. That is, each reference sequence will correspond to a transformed feature. The critical issue in this step is how to design an effective pivot sequence selection method. To date, existing pattern-based methods typically utilize some simple criteria to conduct the reference sequence selection task. For example, those methods based on frequent subsequences use the minimal support constraint as the criterion for reference sequence selection. Apparently, many new and interesting pivot sequence selection methods remain unexplored under our framework. In the subsequent paragraphs of this subsection, we will list some commonly used criteria for selecting reference sequences from the set of candidate pivot sequences.

\begin{mydeff}
($Gap\ constraint$~\cite{deng2010occurrence}). Given two sequences $s = \left \langle s_1,s_2,...,s_l \right \rangle$ and $t = \left \langle t_1,t_2,...,t_r \right \rangle$, if $t$ is the subsequence of $s$ such that $t_1 = s_{i_1},t_2 = s_{i_2}, ... ,t_r = s_{i_r}$, the $gap$ between $i_k$ and $i_{k+1}$ is defined as $Gap(s,i_k,i_{k+1})=i_{k+1}-i_k-1$. Given two thresholds $mingap$ and $maxgap$ ($0 \leq mingap \leq maxgap$), if $mingap \leq Gap(s,i_k,i_{k+1}) \leq maxgap$ ($1 \leq k \leq r-1$), then the occurrence of $t$ in $s$ fulfills the $gap\ constraint$.
\end{mydeff}

\begin{mydeff}
($Minsup\ constraint$~\cite{deng2009contrasting}). Given a set of sequences $D_{c_i}$ with the class label $c_i$ and a sequence $t$, $count_{D_{c_i}}(t)$ is used to denote the number of sequences in $D_{c_i}$ that contain $t$ as a subsequence. The $support$ of $t$ in $D_{c_i}$ is defined as $sup_{D_{c_i}}(t)= \frac{count_{D_{c_i}}(t)}{|D_{c_i}|}$. Given a positive threshold $minsup$, if $sup_{D_{c_i}}(t) \ge minsup$, then $t$ satisfies the $minsup\ constraint$ and $t$ is a frequent sequential pattern in $D_{c_i}$.
\end{mydeff}

\begin{mydeff}
($Mindisc\ constraint$~\cite{liu2014discriminative}). Given two class labels $c_1$ and $c_2$, a sequence $t$ is said to be a discriminative pattern if it is over-expressed on $D_{c_1}$ against $D_{c_2}$ (or the vice versa). To evaluate the discriminative power, many measures/functions have been proposed in the literature~\cite{liu2014discriminative}. If the discriminative function value of $t$ can pass certain constraints, then it satisfies the $mindisc\ constraint$. Here we just list some measures that have been used for selecting discriminative patterns in sequence classification.

\begin{itemize}
\item Discriminative Function (DF) 1~\cite{deng2009contrasting}:
\end{itemize}
\begin{equation}
\begin{gathered}
sup_{D_{c_1}}(t) > minsup, \\
sup_{D_{c_2}}(t) \leq minsup,
\end{gathered}
\end{equation}
where $minsup$ is a given $support$ threshold.

\begin{itemize}
\item Discriminative Function (DF) 2~\cite{deng2010occurrence}:
\end{itemize}
\begin{equation}
\begin{gathered}
occ_{D_{c_1}}(t) > mincount, \\
occ_{D_{c_2}}(t) \leq mincount,
\end{gathered}
\end{equation}
where $occ_{D_{c_1}}(t)= \frac{occount_{D_{c_1}}(t)}{|D_{c_1}|}$ and $mincount$ is a given threshold. The $occount_{D_{c_1}}(t)$ is the number of non-overlapping occurrences of $t$ in $D_{c_1}$.
\begin{itemize}
\item Discriminative Function (DF) 3~\cite{deng2009contrasting}:
\end{itemize}
\begin{equation}
\begin{gathered}
supdi\emph{ff}= sup_{D_{c_1}}(t)-sup_{D_{c_2}}(t).
\end{gathered}
\end{equation}

\begin{itemize}
\item Discriminative Function (DF) 4~\cite{deng2010occurrence}:
\end{itemize}
\begin{equation}
\begin{gathered}
F$-$ratio= \frac{Occ_{between}}{Occ_{within}},
\end{gathered}
\end{equation}
where
\begin{align*}
Occ_{between} ={} & |D_{c_1}|(occ_{D_{c_1}}(t)-\frac{occ_{D_{c_1}}(t)+occ_{D_{c_2}}(t)}{2})^2\\
{} & + |D_{c_2}|(occ_{D_{c_2}}(t)-\frac{occ_{D_{c_1}}(t)+occ_{D_{c_2}}(t)}{2})^2,
\end{align*}
and $Occ_{within}$ is defined as:
\begin{align*}
Occ_{within} ={} & \sum_{j=1}^{|D_{c_1}|}(occount_{D_{c_{1j}}}(t)-occ_{D_{c_1}}(t))^2\\
{} & +\sum_{j=1}^{|D_{c_2}|}(occount_{D_{c_{2j}}}(t)-occ_{D_{c_2}}(t))^2.
\end{align*}

\begin{itemize}
\item Discriminative Function (DF) 5~\cite{he2018mining}:
\end{itemize}
\begin{equation}
\begin{gathered}
GR(t,c_1,c_2) \ge minGR \\
Sig_{con}(t,c_1,c_2) \ge minSig,
\end{gathered}
\end{equation}
where $GR(t,c_1,c_2)=\frac{sup_{c_1}(t)}{sup_{c_2}(t)}$ is the $GrowthRate$ of $t$, $minGR$ is a given $GrowthRate$ threshold. $Sig_{con}(t,c_1,c_2)=min_{q \in Q}\left\{ \frac{GR(t,c_1,c_2)}{GR(q,c_1,c_2)} \right\}$ is used to describe the conditional redundancy, where $Q$ is the set of discriminative sub-patterns of $t$, $minSig$ is a given threshold.

\begin{itemize}
\item Discriminative Function (DF) 6~\cite{lesh1999mining}:
\end{itemize}
\quad \, The chi-squared test is used as the discriminative function to check if the candidate sequence is correlated with at least one class that it is frequent in.
\end{mydeff}

\begin{mydeff}
($Uniqueness\ constraint$~\cite{deng2010occurrence}). A sequence is said to satisfy the $uniqueness\ constraint$ if all its items are unique.
\end{mydeff}

\begin{mydeff}
($Closedness\ constraint$~\cite{tsai2015pso}). A sequence $t$ is said to satisfy the $closedness\ constraint$ if no sequences that contain $t$ as a subsequence have the same $support$ as $t$.
\end{mydeff}

\begin{mydeff}
($Redundancy\ constraint$~\cite{lesh1999mining}).
A sequence $t$ is said to satisfy the $redundancy\ constraint$ if $con\emph{f}(t) \geq \frac{|D_{c_i}|}{|D|}$, where $con\emph{f}(t)=\frac{count_{D_{c_i}}(t)}{count_D(t)}$ is the $con\emph{f}idence$ of $t$.
\end{mydeff}

\begin{mydeff}
($Interestingness\ constraint$~\cite{zhou2016pattern}). Given a set of sequences $D_{c_i}$ with class label $c_i$, two sequences $s = \left \langle s_1,s_2,...,s_l \right \rangle$ and $t = \left \langle t_1,t_2,...,t_r \right \rangle$, if $t$ is the subsequence of $s$ such that $t_1 = s_{i_1},t_2 = s_{i_2}, ... ,t_r = s_{i_r}$, $I_{c_i}(t)=sup_{D_{c_i}}(t) \times C_{c_i}(t)$ is used to denote the $interestingness$ of $t$, where $C_{c_i}(t)= \frac{|t|}{\overline{W_{c_i}}(t)}$ is the $cohesion$ of $t$ in $D_{c_i}(t)$, $\overline{W_{c_i}}(t)= \frac{\sum_{s \in D_{c_i}(t)}W(t,s)}{count_{D_{c_i}}(t)}$ and $W(t,s)=min \left\{i_r-i_1+1 | i_1\leq i_r\right\}$. And the $cohesion$ of $t$ in a sequence $s$ is $C(t,s)=\frac{|t|}{W(t,s)}$. Given two thresholds $minsup$ and $minint$, if $sup_{D_{c_i}}(t) \ge minsup$ and $I_{c_i}(t) \ge minint$, then $t$ satisfies the $interestingness\ constraint$.
\end{mydeff}

\begin{mydeff}
($Level\ constraint$~\cite{egho2017user}).
Given a sequence $t$ and a set of sequences $D$ with $j$ classes, a sequential classification rule $\pi$ is denoted as $\pi:t \to count_{D_{c_1}}(t),count_{D_{c_2}}(t),...,count_{D_{c_j}}(t)$, where $t$ is the body of the rule. From a Bayesian point of view, to choose the best rule is equivalent to maximizing $p(\pi|D)=\frac{p(\pi,D)}{p(D)}=\frac{p(\pi)\times p(D|\pi)}{p(D)}$, where $p(D)$ is a constant, $cost(\pi)=-\log(p(\pi)\times p(D|\pi))$ is used as the evaluation criterion, and
the normalized criterion $level$ is defined as $level(\pi)=1-\frac{cost(\pi)}{cost(\pi_{\emptyset})}$, in which $cost(\pi_{\emptyset})$ is the cost of the null model when the sequence body is empty. If $0<level(\pi)\leq 1$, then $t$ satisfies the $level\ constraint$.
\end{mydeff}

\subsection{Feature Value Generation}
\label{sec:3.2}
\begin{figure*}[htbp]%%图
	\centering  %插入的图片居中表示
	\includegraphics[width=0.9\textwidth]{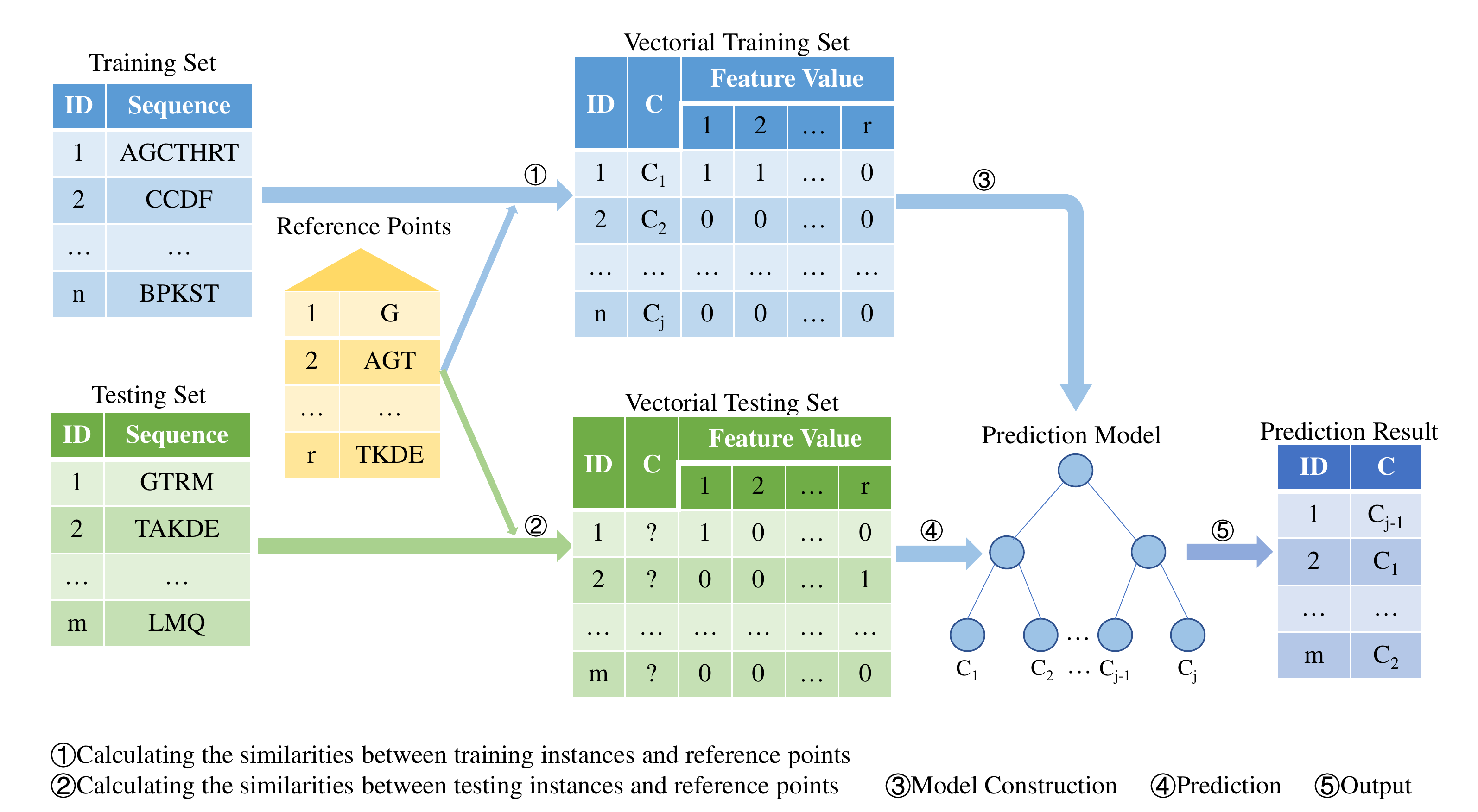}  %插入的图，包括JPG,PNG,PDF,EPS 等，放在源文件目录下
	\caption{The process of feature value generation, model construction and prediction}  % 图片的名称
	\label{fig:mcmthesis-logo3}   %标签，用作引用
\end{figure*}

In the second stage of the presented framework, a similarity function is used to generate vectorial representations for all sequences in both training data and testing data. As shown in the left part of Fig. \ref{fig:mcmthesis-logo3}, this procedure can be further divided into two steps: (1) calculating the similarities between training instances and reference points; (2) calculating the similarities between testing instances and reference points.

In the first step, we utilize a similarity function to transform $TrainD$ into a vectorial training set $TrainD^\prime$ by calculating the similarity between each sequence in $TrainD$ and every reference point in $R$. Each similarity value will be used as the corresponding feature value. The critical issue in this step is how to choose a suitable similarity function. Note that the selection of the similarity function is arbitrary. In other words, any feasible similarity function can be used in this step. In fact, many existing feature-based methods utilize a boolean function as the similarity function, which outputs 1 as the feature value if the reference point is a subsequence of the target sequence and 0 otherwise.

In the second step, we use the same similarity function to transform $TestD$ into a vectorial testing set $TestD^\prime$. Note that the number of features in the transformed vectorial data set is $|R|$, which is the number of reference points.

The similarity function plays an important role in generating feature values. Accordingly, it will have a great impact on the prediction result. For the purpose of summarizing existing research efforts under our framework with respect to the similarity function, here we list some similarity functions between two sequences $s$ and $t$ that have been deployed in the literature.

\begin{itemize}
\item Similarity Function (SF) 1~\cite{lesh1999mining}:
\end{itemize}
\begin{equation}
Sim(s,t)=
\begin{cases}
1, & \text{if  } t \subseteq s,\\
0, & \text{otherwise}.
\end{cases}
\end{equation}

\begin{itemize}
\item Similarity Function (SF) 2~\cite{deng2009contrasting}:
\end{itemize}
\begin{equation}
Sim(s,t)=
\begin{cases}
1, & \text{if  } \alpha\ and\ t\ are\ similar,\\
0, & \text{otherwise}.
\end{cases}
\end{equation}

In Equation (III.7), $similar$ means $ed(\alpha,t) \leq \gamma \times |t|$ ($|s| \geq |t|$), $ed(\alpha,t)$ is the $edit\ distance$ between $\alpha$ and $t$ (the minimum number of operations needed to transform $\alpha$ into $t$, where an operation can be the insertion, deletion, or substitution of a single item), $\alpha$ is a contiguous subsequence of $s$ with $|t|$ items, which is extracted by using a sliding window of length $|t|$ that starts from the first element of $s$.
If $\alpha$ and $t$ are not $similar$, then the sliding window will be repeatedly shifted one position to the right until $|s|-|t|+1$ subsequences have been checked or a new subsequence $\alpha$ $similar$ to $t$ is encountered. $\gamma$ is a given $maximum\ di\emph{ff}erence$ threshold.

\begin{itemize}
\item Similarity Function (SF) 3~\cite{zhou2016pattern}:
\end{itemize}
\begin{equation}
Sim(s,t)=
\begin{cases}
C(t,s), & \text{if  } t \subseteq s,\\
0, & \text{otherwise},
\end{cases}
\end{equation}
where $C(t,s)$ is the $cohesion$ of $t$ in the sequence $s$.

\begin{itemize}
\item Similarity Function (SF) 4~\cite{ntagiou2017protein}:
\end{itemize}
\begin{equation}
Sim(s,t)=
\begin{cases}
occnum, & \text{if  } t \subseteq s,\\
0, & \text{otherwise},
\end{cases}
\end{equation}
where $occnum$ is the number of occurrences of $t$ in $s$.

\begin{itemize}
\item Similarity Function (SF) 5~\cite{deng2010occurrence}:
\end{itemize}
\begin{equation}
Sim(s,t)=
\begin{cases}
occount_{s}(t), & \text{if  } t \subseteq s,\\
0, & \text{otherwise},
\end{cases}
\end{equation}
where $occount_{s}(t)$ is the number of non-overlapping occurrences of $t$ in $s$.

\begin{itemize}
\item Similarity Function (SF) 6~\cite{tsai2015pso}:
\end{itemize}
\begin{equation}
Sim(s,t)= \frac{|LCS(s,t)|}{Max\left\{|s|,|t|\right\}},
\end{equation}
where $|LCS(s,t)|$ is the length of the longest common subsequence, $|s|$ and $|t|$ are the length of $s$ and $t$ respectively.

\subsection{Model Construction and Prediction}
\label{sec:3.3}
In the third stage of the presented framework, we construct a prediction model to make predictions. As shown in the right part of Fig. \ref{fig:mcmthesis-logo3}, this procedure can be further divided into three steps: model construction, prediction and classification result generation.

In the first step, an existing vectorial data classification method is used to construct a prediction model from the vectorial training set $TrainD^\prime$ since we have transformed training sequences into feature vectors in the second stage. Numerous classification methods have been designed for classifying feature vectors (e.g. support vector machines and decision trees)~\cite{Cernadas2014Do,zhang2017up}. After training a classifier with $TrainD^\prime$, the prediction model is ready for classifying unknown samples.

In the second step, we forward the vectorial testing set $TestD^\prime$ to the classifier to make predictions.
In the third step, we output the prediction result and compute the classification accuracy by comparing the predicted class labels with the ground-truth labels.

\section{General Framework for Feature-Based Classification}
\label{sec:4}
In this section, we show that many existing feature-based sequence classification algorithms can be reformulated within the presented reference-based framework. The differences between these algorithms mainly lie in the selection of reference points and similarity functions. As summarized in Table \ref{table1}, we can categorize these existing methods according to three criteria: (1) How to construct the candidate set of reference points? (2) How to choose a set of reference points? (3) Which similarity function should be used? Note that the definitions and notations for different constraints and similarity functions have been presented in Section \ref{sec:3.1} and Section \ref{sec:3.2}. From Table \ref{table1}, we have the following observations.

\begin{table*}
\renewcommand\arraystretch{2}
\caption{The categorization of some existing feature-based sequence classification algorithms under our framework}
\label{table1}
\centering
\begin{tabular}{|m{2cm}<{\centering}|m{0.25\textwidth}<{\centering}|m{0.25\textwidth}<{\centering}|m{0.25\textwidth}<{\centering}|}
\hline
Algorithm & Construction of Candidate Reference Point Set & Selection of Reference Points & Selection of Similarity Function\\
\hline
SCIP\cite{zhou2016pattern} & SubTrainD & minsup, minint\ and\ maxsize\ constraints&SF\ 1/3\\
\hline
Ref.\cite{she2003frequent} & SubTrainD & minsup\ constraint&SF\ 1\\
\hline
Ref.\cite{deng2010occurrence} & SubTrainD & uniqueness, gap, mindisc\ (DF\ 2\ and\ 4)\ constraints &SF\ 5\\
\hline
Ref.\cite{deng2009contrasting} & SubTrainD & gap, mindisc\ (DF\ 1\ and\ 3)\ constraints &SF\ 2\\
\hline
MiSeRe\cite{egho2017user} & SubTrainD & level constraint&SF\ 1\\
\hline
Ref.\cite{ntagiou2017protein} & SubTrainD & minsup\ and\ gap\ constraints& SF\ 1/4\\
\hline
PSO-AB\cite{tsai2015pso} & SubTrainD & minsup\ and\ closeness\ constraints& SF\ 6\\
\hline
FeatureMine\cite{lesh1999mining} & SubTrainD & minsup, redundancy\ and\ mindisc\ (DF\ 6)\ constraints&SF\ 1\\
\hline
CDSPM\cite{he2018mining} & SubTrainD & minsup\ and\ mindisc\ (DF\ 5)\ constraints&SF\ 1\\
\hline
\end{tabular}
\end{table*}

First of all, any sequence over the alphabet can be a potential member of the candidate set of reference points $CR$. However, all feature-based sequence classification algorithms in Table \ref{table1} use $SubTrainD$ to construct $CR$ since the idea of using subsequences as features is quite natural with a good interpretability. Although $SubTrainD$ is a finite set, its size is still very large and most sequences in $SubTrainD$ are useless and redundant for classification. Therefore, it is necessary to explore alternative methods for constructing the set of candidate reference points. For instance, we may use all original sequences in $TrainD$ to construct $CR$, so that the size of $CR$ will be greatly reduced and the corresponding features may be more representative.

Second, many sequence selection criteria have been proposed to select $R$ from $CR$, such as $minsup$ and $mindisc$. The main objective of applying these criteria is to select a subset of sequences that can generate good features for building the classifier. However, it is not an easy task to set suitable thresholds for these constraints to produce a set of reference sequences with moderate size. More importantly, most of these constraints are proposed from the literature of sequential pattern mining, which may be only applicable to the selection of reference sequences from $SubTrainD$. In other words, more general reference point selection strategies should be developed.

Last, the most widely used similarity function in Table \ref{table1} is SF 1, which is a boolean function based on whether the reference point is a subsequence of the sequence in $TrainD$. Although some non-boolean functions have been used, the potential of utilizing more elaborate similarity functions between two sequences still needs further investigation.

Overall, our reference-based sequence classification framework is quite generic, in which many existing pattern-based sequence classification methods can be reformulated as its special variants. Meanwhile, there are still many limitations in current research efforts under this framework. Hence, new and effective sequence classification methods should be developed towards this direction.

\section{New Variants under the Framework}
\label{sec:5}
In addition to encompassing existing pattern-based methods, this framework can also be used as a general platform to design new feature-based sequence classification methods.

As discussed in Section \ref{sec:4}, there are three key ingredients in our framework: the construction of the candidate reference point set, the selection of reference points and the selection of similarity function. Obviously, we will generate a ``new" sequence classification algorithm based on an unexplored combination of these three components. In view of the fact that the number of possible combinations is quite large, it is infeasible to enumerate all these variants. Instead, we will only present two variants that are quite different from existing algorithms to demonstrate the advantage of this framework.

\subsection{The Use of Training Set as the Candidate Set}
\label{sec:5.1}
With our framework, all previous pattern-based sequence classification methods utilize the set $SubTrainD$ as the candidate reference point set $CR$ in the first step. One limitation of this strategy is that the actual size of $CR$ will be very large. As a result, it poses great challenges for the reference point selection task in the consequent step. To alleviate these issues, we propose to use all original sequences in $TrainD$ to construct the set of candidate reference points. The rationale for this candidate set construction method is based on the following observations.

Firstly, all information given for building the classifier is contained in the original training set. In other words, we will not lose any relevant information for the classification task if $TrainD$ is used as the candidate set of reference sequences. In fact, the widely used candidate set $SubTrainD$ is derived from $TrainD$.

Secondly, even we use all the training sequences in $TrainD$ as the reference points, the transformed vectorial data will be a $|TrainD| \times |TrainD|$ table. That is, the number of features is still no larger than the number of samples. Therefore, we do not need to analyze a HDLSS (high-dimension, low-sample-size) data set during the classification stage. In contrast, the number of features may be much larger than the number of samples in the vectorial data obtained from $SubTrainD$ if the parameters are not properly specified during the reference point selection procedure. In fact, we have tested the performance when all training sequences are used as reference points. The experimental results show that this quite simple idea is able to achieve comparable performance in terms of classification accuracy.

Finally, the same idea has been employed in the literature of time series classification~\cite{iosifidis2013multidimensional,kate2016using}. Its success motivates us to investigate the feasibility and advantage in the context of discrete sequence classification.

\subsection{Two Reference Point Selection Methods}
\label{sec:5.2}
To select reference sequences from $TrainD$, those existing constraints proposed in the context of sequential pattern mining are not applicable. Therefore, we have to develop new algorithms to choose a subset of representative reference sequences from $TrainD$. To this end, two different reference sequence selection methods are presented. The first one is an unsupervised method, which selects reference sequences based on cluster analysis without considering the class label information. The second one is a supervised method, which evaluates each candidate sequence according to its discriminative ability across different classes. In the following two sub-sections, we will present the details of these two reference point selection algorithms.

\subsubsection{Unsupervised Reference Point Selection}
\label{sec:5.2.1}
As we have discussed in Section \ref{sec:5.1}, we may choose all sequences in the training set as reference points. However, the number of features in the transformed vectorial data can still be very large if the number of training instances is large. The selection of a small subset of representative training sequences as reference points will greatly reduce the computational burden in the subsequent stage. One natural idea is to divide the training sequences in $CR$ into different clusters using a clustering algorithm~\cite{jain1999data}. Then, we can select a representative sequence from each cluster as the reference point.

To date, many algorithms have been presented for clustering discrete sequences (e.g.~\cite{society2014novel}). We can just adopt an existing sequence clustering algorithm in our pipeline. Here we choose the Group-average Agglomerative Hierarchical Clustering (GAHC) algorithm~\cite{willett1988recent} to fulfill the sequence clustering task. This algorithm is used because it can often generate a high-quality clustering result and can handle any forms of similarity measure.

In the following, we will describe the details of the reference point selection method based on GAHC.

In the first stage, the $i$-th sequence in $CR$ will form a cluster $C_i$.

In the second stage, a similarity function is used to calculate the similarity between each pair of clusters to produce a similarity matrix $Sim$, where $Sim[i,j]$ is the similarity between the two clusters $C_i$ and $C_j$. Many similarity measures have been presented for sequential data (e.g.~\cite{rieck2008linear}). Here we choose the Jaccard coefficient. More specific details on the similarity function will be discussed in Section \ref{sec:5.3}.

In the third stage, we first search the similarity matrix $Sim$ to identify the maximum value $maxSim$, which corresponds to the most similar pair of clusters $C_k$ and $C_l$. Then, these two clusters are merged to form a new cluster $C_k$ and the number of clusters in total is decreased by 1. Meanwhile, the entries related to $C_l$ in $Sim$ are set to be 0 and $Sim$ is updated by recalculating the similarity between $C_k$ and each of the remaining clusters. The similarity between the newly generated cluster and each of the remaining clusters is calculated as the average similarity between all members in the two clusters since we use the $group$-$average$ method. We repeat the third stage until the number of clusters is equal to the number of reference points we want to select.

In the last stage, we select a representative sequence from each cluster. For each cluster, any sequence in this cluster can be used as a representative. To provide a consistent and deterministic output, we use the sequence with the minimum subscript in the cluster as the reference point.

\subsubsection{Supervised Reference Point Selection}
\label{sec:5.2.2}
To choose a subset of representative reference sequences from $TrainD$, we can also employ a supervised method in which the class label information is utilized. As we have discussed in Section \ref{sec:4}, different $mindisc$ constraints have been widely used to evaluate the discriminative power of sequential patterns. Unfortunately, these constraints are only applicable to the selection of reference points from $SubTrainD$. In addition, it is not an easy task to set suitable thresholds to control the number of selected reference points. In order to overcome these limitations, we present a reference point selection method based on hypothesis testing, in which the statistical significance in terms of $p$-value is used to assess the discriminative power of each candidate sequence.

Hypothesis testing is a commonly used method in statistical inference. The usual line of reasoning is as follows: first, formulate the null hypothesis and the alternative hypothesis; second, select an appropriate test statistic; third, set a significance level threshold; finally, reject the null hypothesis if and only if the $p$-value is less than the significance level threshold, where the $p$-value is the probability of getting a value of the
test statistic that is at least as extreme as what is actually observed on condition that the null hypothesis is true.

In order to assess the discriminative power of each candidate sequence in terms of $p$-value, we can use the null hypothesis that this sequence does not belong to any class and all sequences from different classes are drawn from the same population. If the above null hypothesis is true, then the similarities between the candidate sequence and training sequences are drawn from the same population. Therefore, we can formulate the corresponding hypothesis testing problem as a two-sample testing problem~\cite{gibbons2011nonparametric}, where one sample is the set of similarities between the candidate sequence and the training sequences from one target class and another sample is the set of similarities between the candidate sequence and the training sequences from the remaining classes.

Since we test all candidate sequences in $CR$ at the same time, it is actually a multiple hypothesis testing problem.
If no multiple testing correction is conducted, then the number of false positives among reported reference sequences may be very high. To tackle this problem, we adopt the BH procedure to control the FDR (False Discovery Rate)~\cite{benjamini1995controlling}, which is the expected proportion of false positives among all reported sequences.

The reference point selection method based on MHT (Multiple Hypothesis Testing) is shown in Algorithm 1. In the following, we will elaborate on this algorithm in detail.

\begin{algorithm}
%\small
    \caption{Reference Point Selection Based on MHT}
        \begin{algorithmic}[1] %每行显示行号
            \Require Candidate reference sequence set $CR$, $\alpha$
            \Ensure Reference point set $R$
            %\Function {MergerSort}{$Array, left, right$}
                \State $R \leftarrow \emptyset$;
                \For {each $D_{c_i}$ in $CR$}
                    \State $D_+ \leftarrow D_{c_i}$;
                    \State $D_- \leftarrow CR-D_{c_i}$;
                    \For {each sequence $S_k$ in $D_+$}
                         \State $Sim_+\leftarrow \emptyset$;
                         \State $Sim_-\leftarrow \emptyset$;
                         \For{each sequence $S_j$ in $D_+$}
                            \State $calculate$ $Sim[k,j]$;
                            \State $Sim_+ \leftarrow Sim_+ \cup \left\{Sim[k,j] \right\}$;
                         \EndFor
                         \For{each sequence $S_j$ in $D_-$}
                            \State $calculate$ $Sim[k,j]$;
                            \State $Sim_- \leftarrow Sim_- \cup \left\{Sim[k,j] \right\}$;
                         \EndFor
                         \State $S_k.pvalue \leftarrow Utest(Sim_+,Sim_-)$;
                    \EndFor
                    \State $sort$ $D_+$;
                    \State $maxindex \leftarrow 0$;
                    \For {each sequence $S_k$ in $D_+$}
                        \If {$S_k.pvalue \leq \alpha \frac{k}{|D_+|} $}
                            \State $maxindex \leftarrow k$;
                        \EndIf
                    \EndFor
                    \For {$k \leftarrow maxindex+1\ to\ |D_+|-1$}
                        \State $D_+ \leftarrow D_+- \left\{S_k \right\}$;
                    \EndFor
                    \State $R \leftarrow R \cup D_+$;
                \EndFor
                \State \Return{$R$};
        \end{algorithmic}
\end{algorithm}

In the first stage (step 1-4), we select a set of sequences $D_{c_i}$ with the class label $c_i$ from $CR$, then
we regard $D_{c_i}$ as the positive data set $D_+$ and use the set of all remaining sequences in $CR$ as the negative data set $D_-$.

In the second stage (step 5-17), for each sequence $S_k$ in $D_+$, a similarity function is used to calculate the similarity between $S_k$ and each sequence in $D_+$ and $D_-$, where the similarity function is the same as that used in Section \ref{sec:5.2.1} and $Sim[k,j]$ is the similarity between the two sequences $S_k$ and $S_j$. Then, the Mann-Whitney U test~\cite{mann1947test} is used to calculate the $p$-value based on the two similarity set $Sim_+$ and $Sim_-$.

In the third stage (step 18-27), the BH method first sorts sequences in $D_+$ according to their corresponding $p$-value in an ascending order, i.e., $D_+ = \left\{ S_1,S_2,...,S_{|D_+|} \right\}$ ($S_1.pvalue \leq S_2.pvalue \leq ...\leq S_{|D_+|}.pvalue$). Then, we sequentially search $D_+$ to identify the maximal sequence index $maxindex$ which satisfies the condition that $S_k.pvalue \leq \alpha \frac{k}{|D_+|} $, where $\alpha$ is the significance level threshold. Those sequences whose indices are larger than $maxindex$ will be removed from $D_+$.

In the last stage (step 28-30), we select all sequences from $D_+$ as reference points. The whole process will be terminated after each set of sequences from every class has been regarded as $D_+$.

\subsection{Similarity Function}
\label{sec:5.3}
In order to measure the similarity between two sequences, we choose the Jaccard coefficient as the similarity function in our method. The larger the Jaccard coefficient between the two sequences is, the more similar they are.

Given two sequences $s = \left \langle s_1,s_2,...,s_l \right \rangle$ and $t = \left \langle t_1,t_2,...,t_r \right \rangle$, the Jaccard coefficient is defined as:
\begin{equation}
    J(s,t) = \frac{|s \cap t|}{|s|+|t|-|s \cap t|},
\end{equation}
where $|s \cap t|$ is the number of items in the intersection of $s$ and $t$. However, this may lose the order information of sequences. To alleviate this issue, we use the LCS (Longest Common Subsequence) between $s$ and $t$ to replace $s \cap t$. Then, the Jaccard coefficient is redefined as:
\begin{equation}
		J(s,t) = \frac{|LCS(s,t)|}{|s|+|t|-|LCS(s,t)|}.
\end{equation}
\begin{mydefff}
Given two sequences $s = \left \langle a, b, c, d, e\right \rangle$ and $t = \left \langle e, c, d, c\right \rangle$, the $LCS(s,t)$ is $\left \langle c, d\right \rangle$, then the modified Jaccard coefficient is
\begin{align*}
J(abcde,ecdc) & =  \frac{2}{5+4-2} \approx  0.286.
\end{align*}
\end{mydefff}

Note that we can also use other similarity functions in the literature, such as those methods summarized and reviewed in~\cite{rieck2008linear}. The choice of a more appropriate similarity function may yield better performance than the modified Jaccard coefficient. In order to check the effect of similarity function on the classification performance, we also consider the following two alternative similarity functions.

The first one is the String Subsequence Kernel (SSK)~\cite{lodhi2002text}. The main idea of SSK is to compare two sequences by means of the subsequences they contain in common. That is, the more subsequences in common, the more similar they are.

Given two sequences $s = \left \langle s_1,s_2,...,s_l \right \rangle$ and $t = \left \langle t_1,t_2,...,t_r \right \rangle$ and a parameter $n$, the SSK is defined as:
\begin{equation}
\begin{aligned}
	K_n(s,t) & =  \langle \Phi(s), \Phi(t) \rangle \\
				& =  \sum_{u \in I^n} \phi_u(s).\phi_u(t) \\
				& =  \sum_{u \in I^n} \sum_{u\subseteq s} \lambda^{l_s(u)}
			\sum_{u\subseteq t} \lambda^{l_t(u)} \\
			& =  \sum_{u \in I^n} \sum_{u\subseteq s}  \sum_{u\subseteq t} \lambda^{l_s(u) + l_t(u)},
\end{aligned}
\end{equation}
where $\phi_u(s)$ is the feature mapping for the sequence $s$ and each $u\in I^n$, $I$ is a finite alphabet, $I^n$ is the set of all subsequences of length $n$ and $u$ is a subsequence of $s$ such that $u_1 = s_{i_1},u_2 = s_{i_2}, ... ,u_n = s_{i_n}$, $l_s(u) = i_n-i_1+1$ is the length of $u$ in $s$, $\lambda \in (0,1)$ is a decay factor which is used to penalize the gap. The calculation steps are as follows: enumerate all subsequences of length $n$, compute the feature vectors for the two sequences, and then compute the similarity. The normalized kernel value is given by
\begin{equation}
		\hat{K}_n(s,t) = \frac{K_n(s,t)}{\sqrt{K_n(s,s) K_n(t,t)}}.
\end{equation}
\begin{mydefff}
Given two sequences $s = \left \langle a, b, c, d, e\right \rangle$ and $t = \left \langle e, c, d, c\right \rangle$, the subsequences of length 1 ($n$=1) are $a,b,c,d,e$. The corresponding feature vector for each of the sequences can be denoted as $\phi_1(s) = \left \langle \lambda,\lambda,\lambda,\lambda,\lambda\right \rangle$ and $\phi_1(t) = \left \langle 0,0,2\lambda,\lambda,\lambda\right \rangle$, then the normalized kernel value is
\begin{align*}
\hat{K}_1(abcde,ecdc) & =  \frac{K_1(abcde,ecdc)}{\sqrt{K_1(abcde,abcde) K_1(ecdc,ecdc)}}\\
			 & =  \frac{4\lambda^2}{\sqrt{5\lambda^2 \times 6\lambda^2}} \approx  0.73.
\end{align*}
\end{mydefff}

When this function is employed in our method, $n$ = 1 is used as the default parameter setting. Although the setting of $n$ = 1 may lose the order information, it will greatly reduce the computational cost and can provide satisfactory results in practice.

Another alternative similarity function is the normalized LCS. The larger the normalized LCS between two sequences is, the more similar they are.

Given two sequences $s = \left \langle s_1,s_2,...,s_l \right \rangle$ and $t = \left \langle t_1,t_2,...,t_r \right \rangle$, the normalized LCS is defined as:
\begin{equation}
Sim(s,t)= \frac{|LCS(s,t)|}{Min\left\{|s|,|t|\right\}},
\end{equation}
where $|LCS(s,t)|$ is the length of the longest common subsequence, $|s|$ is length of $s$, and $|t|$ is the length of $t$.
\begin{mydefff}
Given two sequences $s = \left \langle a, b, c, d, e\right \rangle$ and $t = \left \langle e, c, d, c\right \rangle$, the $LCS(s,t)$ is $\left \langle c, d\right \rangle$, then the normalized LCS is
\begin{align*}
Sim(abcde,ecdc) & =  \frac{2}{4} =  0.5.
\end{align*}
\end{mydefff}

\section{Experiments}
\label{sec:6}
To demonstrate the feasibility and advantages of this new framework, we conducted experiments on fourteen real sequential data sets. We compared our two algorithms derived under the reference-based framework with other sequence classification algorithms in terms of classification accuracy. All experiments were conducted on a PC with Intel(R) Xeon(R) CPU 2.40GHz and 12G Memory. All the reported accuracies in the experiments were the average accuracies obtained by repeating the 5-fold cross-validation 5 times except SCIP (accuracies in SCIP were obtained using 10-fold cross-validation because this is a fixed setting in software package provided by the author).

\begin{table}
%\small
\caption{Summary of the Sequential Data Sets Used in the Experiments}
\label{table2}
\centering
\begin{tabular}{ccccccc}
\toprule
Dataset &  $|D|$ &  \#items & minl & maxl& avgl &  \#classes \\
\midrule
  Activity &  35   &  10   &  12   &   43   &  21.14    &  2   \\
  Aslbu &  424   &  250   &  2   &   54   &   13.05   &  7   \\
  Auslan2 &  200   &  16   &  2   &  18    &  5.53    &  10   \\
  Context &  240   &  94   &  22   &  246    &  88.39    & 5    \\
  Epitope &  2392   &  20   & 9    &  21    &  15    &  2   \\
  Gene &  2942   &  5   &  41   &  216    &  86.53    &  2   \\
  News &  4976   &  27884   &  1   & 6779     &  139.96    &  5   \\
  Pioneer &  160   & 178    & 4    &  100    &  40.14   &  3   \\
  Question &   1731  &  3612   &  4   &  29    &  10.17    &  2   \\
  Reuters &  1010   &  6380   &  4   &  533    &  93.84    &  4   \\
  Robot &  4302   &  95  &  24   &  24    &  24    &  2   \\
  Skating &  530   &  82   & 18    & 240     &   48.12   & 7    \\
  Unix &   5472  &  1697   &   1  &  1400    &  32.34    &  4   \\
  Webkb &   3667  &  7736   &  1   &  20628    &  129.37    & 3    \\
\bottomrule
\end{tabular}
\end{table}

\subsection{Data Sets}
\label{sec:6.1}
We choose fourteen benchmark data sets which are widely used for evaluating sequence classification algorithms: Activity~\cite{lichman2013uci}, Aslbu~\cite{fradkin2015mining}, Auslan2~\cite{fradkin2015mining}, Context~\cite{mantyjarvi2004sensor}, Epitope~\cite{deng2009contrasting}, Gene~\cite{wei2014improved}, News~\cite{zhou2016pattern}, Pioneer~\cite{fradkin2015mining}, Question~\cite{kim2014convolutional}, Reuters~\cite{zhou2016pattern}, Robot~\cite{zhou2016pattern}, Skating~\cite{fradkin2015mining}, Unix~\cite{zhou2016pattern}, Webkb~\cite{zhou2016pattern}. The main characteristics of these data sets are summarized in Table \ref{table2}, where $|D|$ represents the number of sequences in the data set, \#items denotes the number of distinct elements, minl, maxl and avgl are used to denote the minimum length, maximum length and average length of the sequences respectively, and \#classes represents the number of distinct classes in the data set.

\begin{table*}
%\small
\caption{Performance comparison of different algorithms in terms of the classification accuracy}
\label{table3}
\centering
\begin{tabular}{c|ccccccccc|cc}
\toprule
Dataset & Classifier & R-A      & R-MHT  & R-GAHC & MiSeRe & FSP    & DSP   & Sqn2VecSEP & Sqn2VecSIM&Classifier&SCIP\\
\midrule
     &        NB     &  0.966  & 0.977  & 0.811  & 1.000      & 0.960   & 0.960    & 1.000  & 1.000 &SCII\_HAR &0.663\\
     &        DT    &  0.931  & 0.931  & 0.794  & 0.960   & 1.000      & 1.000       & 0.900  & 0.800 &SCII\_MA&0.675\\
Activity &    SVM    &  0.977  & 0.926  & 0.629  & 1.000      & 0.994 & 0.994  & 1.000  & 0.950 &SCIS\_HAR&0.967\\
     &        KNN    &  0.983  & 0.811  & 0.800  & 1.000      & 0.886 & 0.886  & 1.000  & 0.950 &SCIS\_MA&1.000\\
     &               &        &        &        &        &       &        &    &      &        & \\
          &   NB     &  0.574  & 0.561  & 0.449  & 0.548 & 0.527 & 0.420  & 0.298  & 0.554 &SCII\_HAR &0.540\\
     &        DT    &  0.523  & 0.527  & 0.480  & 0.565 & 0.542 & 0.459  & 0.405  & 0.484 &SCII\_MA&0.526\\
Aslbu &       SVM    &  0.638  & 0.625  & 0.483  & 0.571 & 0.581 & 0.455  & 0.498  & 0.633 &SCIS\_HAR&0.553\\
     &        KNN    &  0.563  & 0.310  & 0.479  & 0.574 & 0.531  & 0.464  & 0.544  & 0.591 &SCIS\_MA&0.536\\
     &               &        &        &        &        &       &        &    &      &        & \\
          &   NB     &  0.322   & 0.330   & 0.334   & 0.304  & 0.292  & 0.145   & 0.260  & 0.290 &SCII\_HAR &0.100\\
     &        DT    &  0.317   & 0.308   & 0.308   & 0.314  & 0.330   & 0.170    & 0.270  & 0.270 &SCII\_MA&0.095\\
Auslan2 &     SVM    &  0.328     & 0.323   & 0.326   & 0.300    & 0.318  & 0.170    & 0.290  & 0.310 &SCIS\_HAR&0.200\\
     &        KNN    &  0.327    & 0.304   & 0.333   & 0.302  & 0.309  & 0.166   & 0.310  & 0.200 &SCIS\_MA&0.175\\
     &               &        &        &        &        &       &        &    &      &        & \\
          &   NB     &  0.780   & 0.778   & 0.772  & 0.938 & 0.812 & 0.572  & 0.900  & 0.900 &SCII\_HAR &0.613\\
     &        DT    &  0.791  & 0.798  & 0.745  & 0.841 & 0.873 & 0.575  & 0.600  & 0.542 &SCII\_MA&0.617\\
Context &     SVM    &  0.939  & 0.937  & 0.755  & 0.927 & 0.868 & 0.577  & 0.933  & 0.900 &SCIS\_HAR&0.796\\
     &        KNN    &  0.871  & 0.853  & 0.813  & 0.896 & 0.839 & 0.585  & 0.900  & 0.858 &SCIS\_MA&0.867\\
     &               &        &        &        &        &       &        &    &      &        & \\
          &   NB     &  0.675  & 0.663  & 0.751  & 0.588 & 0.696 & 0.671  & 0.779  & 0.761 &SCII\_HAR &0.684\\
     &        DT    &  0.839  & 0.842  & 0.815  & 0.842 & 0.814 & 0.750  & 0.813  & 0.800 &SCII\_MA&0.712\\
Epitope &     SVM    &  0.855  & 0.838  & 0.769  & 0.834 & 0.758 & 0.716  & 0.802  & 0.800 &SCIS\_HAR&0.705\\
     &        KNN    &  0.932   & 0.925  & 0.918  & 0.924 & 0.898 & 0.801  & 0.863  & 0.841 &SCIS\_MA&0.721\\
     &               &        &        &        &        &       &        &    &      &        & \\
          &   NB     &  0.998  & 0.997  & 1.000  & 1.000      & 1.000      & 1.000       & 1.000  & 1.000 &SCII\_HAR &1.000\\
     &        DT    &  0.997  & 0.997  & 0.998  & 1.000      & 1.000 & 1.000  & 0.967  & 0.976 &SCII\_MA&1.000\\
Gene &        SVM    &  1.000       & 1.000       & 1.000       & 1.000      & 1.000      & 1.000       & 0.999  & 1.000 &SCIS\_HAR&1.000\\
     &        KNN    &  1.000       & 1.000       & 1.000       & 1.000      & 1.000      & 1.000       & 1.000  & 1.000 &SCIS\_MA&1.000\\
     &               &        &        &        &        &       &        &    &      &        & \\
          &   NB     &  0.754  & 0.765  & 0.645  & 0.866 & 0.595 & 0.714  & 0.946  & 0.967 &SCII\_HAR &0.936\\
     &        DT    &  0.719  & 0.724   & 0.654  & 0.837 & 0.793 & 0.767  & 0.600  & 0.555 &SCII\_MA&0.942\\
News &        SVM    &  0.975  & 0.969  & 0.874  & 0.905  & 0.771 & 0.775  & 0.972  & 0.974 &SCIS\_HAR&0.910\\
     &        KNN    &  0.856  & 0.350  & 0.756  & 0.732 & 0.544 & 0.761  & 0.918  & 0.912 &SCIS\_MA&0.918\\
     &               &        &        &        &        &       &        &    &      &        & \\
          &   NB     &  0.977  & 0.931  & 0.796  & 0.891  & 0.932 & 0.852  & 0.975  & 0.963 &SCII\_HAR &0.963\\
     &        DT    &  0.883  & 0.870  & 0.826   & 0.988 & 0.964 & 0.926  & 0.788  & 0.825 &SCII\_MA&0.963\\
Pioneer &     SVM    &  0.980  & 0.952  & 0.638  & 0.989 & 0.983 & 0.930  & 0.950  & 0.988 &SCIS\_HAR&0.975\\
     &        KNN    &  0.990  & 0.477  & 0.828  & 0.975 & 0.933 & 0.926  & 0.975  & 0.988 &SCIS\_MA&0.975\\
     &               &        &        &        &        &       &        &    &      &        & \\
          &   NB     &  0.846  & 0.828  & 0.840   & 0.870 & 0.763 & 0.763  & 0.736  & 0.762 &SCII\_HAR &0.833\\
     &        DT    &  0.889  & 0.881  & 0.877  & 0.885 & 0.822 & 0.759  & 0.717  & 0.809 &SCII\_MA&0.785\\
Question &    SVM    &  0.949  & 0.947  & 0.868  & 0.902 & 0.814 & 0.763  & 0.789  & 0.881 &SCIS\_HAR&0.846\\
     &        KNN    &  0.895  & 0.879  & 0.889  & 0.897  & 0.819 & 0.762  & 0.828  & 0.874 &SCIS\_MA&0.837\\
     &               &        &        &        &        &       &        &    &      &        & \\
          &   NB     &  0.892  & 0.893  & 0.808  & 0.903  & 0.831 & 0.765  & 0.921  & 0.905 &SCII\_HAR &0.951\\
     &        DT    &  0.878  & 0.878  & 0.843  & 0.912 & 0.903 & 0.897  & 0.826  & 0.741 &SCII\_MA&0.953\\
Reuters &     SVM    &  0.976   & 0.970  & 0.858  & 0.962 & 0.933 & 0.915  & 0.984  & 0.974 &SCIS\_HAR&0.957\\
     &        KNN    &  0.958  & 0.452  & 0.918  & 0.899 & 0.894 & 0.900  & 0.960  & 0.960 &SCIS\_MA&0.956\\
     &               &        &        &        &        &       &        &    &      &        & \\
          &   NB     &  0.871  & 0.866  & 0.832  & 0.826 & 0.735 & 0.718  & 0.808  & 0.822 &SCII\_HAR &0.795\\
     &        DT    &  0.880  & 0.879    & 0.871  & 0.900 & 0.843 & 0.742  & 0.811  & 0.778 &SCII\_MA&0.822\\
Robot &       SVM    &  0.955  & 0.952  & 0.902  & 0.913 & 0.780 & 0.723  & 0.840  & 0.834 &SCIS\_HAR&0.817\\
     &        KNN    &  0.947  & 0.947  & 0.942  & 0.937 & 0.860 & 0.743  & 0.945  & 0.949 &SCIS\_MA&0.819\\
     &               &        &        &        &        &       &        &    &      &        & \\
          &   NB     &  0.281  & 0.271  & 0.197  & 0.290 & 0.240 & N/A     & 0.336  & 0.321&SCII\_HAR &0.181\\
     &        DT    &  0.258  & 0.215  & 0.204  & 0.258 & 0.272 & N/A     & 0.226  & 0.230 &SCII\_MA&0.181\\
Skating &     SVM    &  0.375  & 0.277   & 0.208  & 0.293 & 0.299 & N/A     & 0.370  & 0.321 &SCIS\_HAR&0.189\\
     &        KNN    &  0.290  & 0.203  & 0.245  & 0.241 & 0.191 & N/A     & 0.302  & 0.340 &SCIS\_MA&0.191\\
     &               &        &        &        &        &       &        &    &      &        & \\
          &   NB     &  0.718  & 0.764  & 0.772  & 0.768 & 0.699 & 0.607  & 0.703  & 0.566 &SCII\_HAR &0.837\\
     &        DT    &  0.887  & 0.883  & 0.874  & 0.899 & 0.819 & 0.750  & 0.776  & 0.756 &SCII\_MA&0.838\\
Unix &        SVM    &  0.927   & 0.921  & 0.906  & 0.915 & 0.820 & 0.748  & 0.899  & 0.872 &SCIS\_HAR&0.857\\
     &        KNN    &  0.869  & 0.822  & 0.865   & 0.873 & 0.803 & 0.745   & 0.892  & 0.891 &SCIS\_MA&0.842\\
     &               &        &        &        &        &       &        &    &      &        & \\
          &   NB     &  0.710  & 0.720  & 0.641   & 0.845 & 0.701 & 0.833  & 0.858  & 0.886 &SCII\_HAR &0.897\\
     &        DT    &  0.820  & 0.821   & 0.788   & 0.874 & 0.869 & 0.862  & 0.629  & 0.635 &SCII\_MA&0.911\\
Webkb &       SVM    &  0.954  & 0.952  & 0.880  & 0.927 & 0.895 & 0.869  & 0.934  & 0.940 &SCIS\_HAR&0.894\\
     &        KNN    &  0.887  & 0.544  & 0.851  & 0.779  & 0.843 & 0.858  & 0.772  & 0.691 &SCIS\_MA&0.901\\
\bottomrule
\end{tabular}
\end{table*}

\begin{table*}
%\small
\caption{The average classification accuracies of different methods over all data sets used in the experiment}
\label{table4}
\centering
\begin{tabular}{ccccccccc|cc}
\toprule
Classifier &  R-A      & R-MHT  & R-GAHC & MiSeRe & FSP    & DSP    &Sqn2VecSEP& Sqn2VecSIM&Classifier&SCIP\\
\midrule
NB         & 0.740 & 0.739 & 0.689  & 0.760  & 0.699    & 0.694 & 0.751   & \textbf{0.764}    &SCII\_HAR  & 0.714\\
DT         & 0.758  & 0.754 & 0.720  & \textbf{0.791}  & 0.775    & 0.743 & 0.666   & 0.657    &SCII\_MA   & 0.716\\
SVM        & \textbf{0.845} & 0.828 & 0.721  & 0.817  & 0.772    & 0.741 & 0.804   & 0.813    &SCIS\_HAR  & 0.762\\
KNN        & \textbf{0.812} & 0.634 & 0.760  & 0.788  & 0.739    & 0.738 & 0.801   & 0.789    &SCIS\_MA   & 0.767\\
\midrule
AVG        & \textbf{0.789} & 0.739 & 0.722  & \textbf{0.789}  & 0.746    & 0.729 & 0.756   & 0.756    &AVG   & 0.740\\
\bottomrule
\end{tabular}
\end{table*}

\begin{table*}
%\small
\caption{The average classification accuracies of different similarity functions over all data sets used in the experiment}
\label{table5}
\centering
\begin{tabular}{cccccccccc}
\toprule
Classifier & R-A-J &  R-A-S  &  R-A-N & R-MHT-J  & R-MHT-S     & R-MHT-N     &R-GAHC-J& R-GAHC-S&R-GAHC-N\\
\midrule
NB         & \textbf{0.740}  & 0.675 & 0.718  & 0.739  & 0.677    & 0.711 & 0.689   & 0.672    &0.674  \\
DT         & \textbf{0.758} & 0.733 & 0.747  & 0.754  & 0.723   & 0.744 & 0.720   & 0.702    & 0.703  \\
SVM        & \textbf{0.845} & 0.793 & 0.837  & 0.828  & 0.779    & 0.829 & 0.721   & 0.746    & 0.762 \\
KNN        & \textbf{0.812} & 0.752 & 0.802  & 0.634  & 0.738   & 0.773 & 0.760   & 0.706    & 0.749  \\
\midrule
AVG        & \textbf{0.789} & 0.738 & 0.776  & 0.739  & 0.730   & 0.764 & 0.722   & 0.706    & 0.722  \\
\bottomrule
\end{tabular}
\end{table*}

\subsection{Parameter Settings}
\label{sec:6.2}
Our two algorithms are denoted by R-MHT (Reference Point Selection Based on MHT) and R-GAHC (Reference Point Selection Based on GAHC), respectively. In addition, the method that uses all sequences in $TrainD$ as reference points is denoted as R-A, which is also included in the performance comparison. We compare our algorithms with five existing sequence classification algorithms: MiSeRe$\footnote{http://www.misere.co.nf}$~\cite{egho2017user}, Sqn2Vec$\footnote{https://github.com/nphdang/Sqn2Vec}$~\cite{nguyen2018sqn2vec}, SCIP$\footnote{http://adrem.ua.ac.be/sites/adrem.ua.ac.be/files/SCIP.zip}$~\cite{zhou2016pattern}, FSP (the algorithm based on frequent sequential patterns) and DSP (the algorithm based on discriminative sequential patterns).

In MiSeRe, $num\_of\_rules$ is specified to be 1024 and $execution\_time$ is set to be 5 minutes for all data sets.

Sqn2Vec is an unsupervised method for learning sequence embeddings from both singleton symbols and sequential patterns. It has two variants: Sqn2VecSEP and Sqn2VecSIM, where Sqn2VecSEP (Sqn2VecSIM) generates sequence representations from singleton symbols and sequential patterns separately (simultaneously). In these two variants, $minsup$ = 0.05, $maxgap$ = 4 and the embedding dimension $d$ is set to be 128 for all data sets.

SCIP is a sequence classification method based on interesting patterns, which has four different variants: SCII\_HAR, SCII\_MA, SCIS\_HAR and SCIS\_MA. In the experiments, the following parameter setting is used in all data sets: $minsup$ = 0.05, $minint$ = 0.02, $maxsize$ = 3, $con\emph{f}$ = 0.5 and $topk$ = 11.

Frequent sequential patterns have been widely used as features in sequence classification. To include the algorithm based on frequent sequential patterns in the comparison (denoted by FSP), we employ the PrefixSpan algorithm~\cite{pei2001prefixspan} as the frequent sequential pattern mining algorithm. The parameters are specified as follows: $maxsize$ = 3 and $minsup$ = 0.3 for all data sets except Context (the $minsup$ in Context is set to be 0.9 in order to avoid the generation of too many patterns).

Similarly, discriminative sequential patterns are widely used as features in many sequence classification algorithms and applications as well. To include the algorithm based on discriminative sequential patterns in the comparison (denoted by DSP), we first use the PrefixSpan algorithm to mine a set of frequent sequential patterns and then detect discriminative patterns from the frequent pattern set. The parameters for PrefixSpan are identical to those used in FSP and $minGR$ = 3 is used as the threshold for filtering discriminative sequential patterns.

\subsection{Results}
\label{sec:6.3}
In Table \ref{table3}, the detailed performance comparison results in terms of classification accuracies are presented. Note that the result of DSP on the Skating data set is N/A because we cannot find any discriminative patterns from this data set based on the given parameter setting. In the experiments, $\alpha$ = 0.05 is used for R-MHT and $pointnum$ is specified to be 1/10 of the size of $TrainD$ for R-GAHC. After transforming sequences into feature vectors, we chose NB (Naive Bayes), DT (Decision Tree), SVM (Support Vector Machine), KNN ($k$ Nearest Neighbors) as the classifiers. The implementation of each classifier was obtained from WEKA~\cite{hall2009weka} except Sqn2Vec. In Sqn2Vec, all classifiers were obtained from scikit-learn~\cite{pedregosa2011scikit} since its source code is written in python.

In order to have a global picture of the overall performance of different algorithms, we calculate the average accuracy over all data sets for each classifier. The corresponding average accuracies for different methods are recorded in Table \ref{table4}. The results show that among our two methods, R-MHT can achieve better performance than R-GAHC when NB, DT and SVM are used as the classifier. However, R-MHT has a bad performance when KNN is used as the classifier. Since we select a representative sequence from each cluster in R-GAHC and any sequence in a cluster can be used as a representative, we may miss the most representative sequence. Meanwhile, the choice of clustering method and the specification of the number of clusters will influence the results. In addition, the R-A method outperforms R-MHT and R-GAHC since we will not lose any relevant information for the classification task when all training sequences are used as reference points. However, the feature dimension will be very high in R-A, which will incur high computational cost in practice.

Compared with other classification methods, our methods are able to achieve comparable performance. In particular, R-A and MiSeRe~\cite{egho2017user} can achieve the highest average classification accuracy among all competitors since all information given for building the classifier is contained in the reference point set in R-A. The reason why R-MHT and R-GAHC are slightly worse may be that their reference points are less distinct from each other in different classes and some sequences that are important for classification are missed. It is quite amazing since R-A is a very simple algorithm derived from our framework. This indicates that the proposed reference-based sequence classification framework is quite useful in practice. It can be expected more accurate feature-based sequence classification methods will be developed under this framework in the future. From Table \ref{table3} and Table \ref{table4}, it can be also observed that none of the algorithms in the comparison can always achieve the best performance across all data sets. Therefore, more research efforts still should be devoted to the development of effective sequence classification algorithms.

The use of different similarity functions may affect the performance of our algorithms. To investigate this issue, we use two additional similarity functions in the experiments for comparison: SSK and the normalized LCS, whose details have been introduced in Section \ref{sec:5.3}.

Table \ref{table5} presents the average classification accuracies of different similarity functions over all data sets. Jaccard coefficient, SSK and normalized LCS are denoted as J, S and N, respectively. In Table \ref{table5}, R-A-J means that the Jaccard coefficient is used as the similarity function in R-A. Other notations in this table can be interpreted in a similar manner. The results show that the use of different similarity functions can affect the performance of our algorithms. Among these three similarity functions, the use of the Jaccard coefficient as the similarity function can achieve better performance in most cases. However, R-MHT-J has unsatisfactory performance when KNN is used as the classifier. It can be also observed that none of the similarity functions is always the best performer. Therefore, more suitable similarity functions should be developed.

The above experimental results and analysis show that the proposed new methods based on our framework can achieve comparable performance to those state-of-the-art sequence classification algorithms, which demonstrate the feasibility and advantages of our framework. And our framework is quite general and flexible since the selection of both reference points and similarity functions is arbitrary. However, since the feature selection and classifier construction in our framework are separate and any existing vectorial data classification methods can be used to tackle the sequence classification problem, some features that are critical to the classifier may be filtered out during the selection process.

\section{Conclusion}
\label{sec:7}
In this paper, we present a reference-based sequence classification framework by generalizing the pattern-based methods. This framework is quite general and flexible, which can be used as a general platform to develop new algorithms for sequence classification. To verify this point, we present several new feature-based sequence classification algorithms under this new framework. A series of comprehensive experiments on real data sets show that our methods are capable of achieving better classification accuracy than existing sequence classification algorithms. Thus, the reference-based sequence classification framework is quite promising and useful in practice.

In future work, we intend to explore more appropriate reference sequence selection methods and similarity functions to improve the performance and reduce the computational cost. As a result, more accurate feature-based sequence classification methods would be derived under this framework.

	\bibliographystyle{IEEEtran}
	\bibliography{ref}

\begin{thebibliography}{10}
\providecommand{\url}[1]{#1}
\csname url@samestyle\endcsname
\providecommand{\newblock}{\relax}
\providecommand{\bibinfo}[2]{#2}
\providecommand{\BIBentrySTDinterwordspacing}{\spaceskip=0pt\relax}
\providecommand{\BIBentryALTinterwordstretchfactor}{4}
\providecommand{\BIBentryALTinterwordspacing}{\spaceskip=\fontdimen2\font plus
\BIBentryALTinterwordstretchfactor\fontdimen3\font minus
  \fontdimen4\font\relax}
\providecommand{\BIBforeignlanguage}[2]{{%
\expandafter\ifx\csname l@#1\endcsname\relax
\typeout{** WARNING: IEEEtran.bst: No hyphenation pattern has been}%
\typeout{** loaded for the language `#1'. Using the pattern for}%
\typeout{** the default language instead.}%
\else
\language=\csname l@#1\endcsname
\fi
#2}}
\providecommand{\BIBdecl}{\relax}
\BIBdecl

\bibitem{han2011data}
J.~Han, J.~Pei, and M.~Kamber, \emph{Data mining: concepts and
  techniques}.\hskip 1em plus 0.5em minus 0.4em\relax Elsevier, 2011.

\bibitem{deshpande2002evaluation}
M.~Deshpande and G.~Karypis, ``Evaluation of techniques for classifying
  biological sequences,'' in \emph{Proceedings of the 6th Paciﬁc-Asia
  Conference on Advances in Knowledge Discovery and Data Mining}.\hskip 1em
  plus 0.5em minus 0.4em\relax Berlin, Germany: Springer, 2002, pp. 417--431.

\bibitem{Xing2010A}
Z.~Xing, J.~Pei, and E.~Keogh, ``A brief survey on sequence classification,''
  \emph{Acm Sigkdd Explorations Newsletter}, vol.~12, no.~1, pp. 40--48, 2010.

\bibitem{Cernadas2014Do}
E.~Cernadas and D.~Amorim, ``Do we need hundreds of classifiers to solve real
  world classification problems?'' \emph{Journal of Machine Learning Research},
  vol.~15, no.~1, pp. 3133--3181, 2014.

\bibitem{zhou2016pattern}
C.~Zhou, B.~Cule, and B.~Goethals, ``Pattern based sequence classification,''
  \emph{IEEE Transactions on Knowledge and Data Engineering}, vol.~28, no.~5,
  pp. 1285--1298, 2016.

\bibitem{exarchos2008two}
T.~P. Exarchos, M.~G. Tsipouras, C.~Papaloukas, and D.~I. Fotiadis, ``A
  two-stage methodology for sequence classification based on sequential pattern
  mining and optimization,'' \emph{Data \& Knowledge Engineering}, vol.~66,
  no.~3, pp. 467--487, 2008.

\bibitem{lo2009classification}
D.~Lo, H.~Cheng, J.~Han, S.-C. Khoo, and C.~Sun, ``Classification of software
  behaviors for failure detection: a discriminative pattern mining approach,''
  in \emph{Proceedings of the 15th ACM SIGKDD International Conference on
  Knowledge Discovery and Data Mining}.\hskip 1em plus 0.5em minus 0.4em\relax
  ACM, 2009, pp. 557--566.

\bibitem{she2003frequent}
R.~She, F.~Chen, K.~Wang, M.~Ester, J.~L. Gardy, and F.~S. Brinkman,
  ``Frequent-subsequence-based prediction of outer membrane proteins,'' in
  \emph{Proceedings of the 9th ACM SIGKDD International Conference on Knowledge
  Discovery and Data Mining}.\hskip 1em plus 0.5em minus 0.4em\relax ACM, 2003,
  pp. 436--445.

\bibitem{hopf2010mining}
T.~Hopf and S.~Kramer, ``Mining class-correlated patterns for sequence
  labeling,'' in \emph{Proceedings of the 2010 Springer International
  Conference on Discovery Science}.\hskip 1em plus 0.5em minus 0.4em\relax
  Springer, 2010, pp. 311--325.

\bibitem{haleem2014novel}
H.~Haleem, P.~K. Sharma, and M.~S. Beg, ``Novel frequent sequential patterns
  based probabilistic model for effective classification of web documents,'' in
  \emph{Proceedings of the 5th International Conference on Computer and
  Communication Technology}.\hskip 1em plus 0.5em minus 0.4em\relax IEEE, 2014,
  pp. 361--371.

\bibitem{deng2010occurrence}
K.~Deng and O.~R. Za{\"\i}ane, ``An occurrence based approach to mine emerging
  sequences,'' in \emph{Proceedings of the 2010 Springer International
  Conference on Data Warehousing and Knowledge Discovery}.\hskip 1em plus 0.5em
  minus 0.4em\relax Berlin, Germany: Springer, 2010, pp. 275--284.

\bibitem{deng2009contrasting}
------, ``Contrasting sequence groups by emerging sequences,'' in
  \emph{Proceedings of the 12th International Conference on Discovery
  Science}.\hskip 1em plus 0.5em minus 0.4em\relax Berlin, Germany: Springer,
  2009, pp. 377--384.

\bibitem{he2019significance}
Z.~He, S.~Zhang, and J.~Wu, ``Significance-based discriminative sequential
  pattern mining,'' \emph{Expert Systems with Applications}, vol. 122, pp.
  54--64, 2019.

\bibitem{fradkin2015mining}
D.~Fradkin and F.~M{\"o}rchen, ``Mining sequential patterns for
  classification,'' \emph{Knowledge and Information Systems}, vol.~45, no.~3,
  pp. 731--749, 2015.

\bibitem{yahyaoui2016feature}
H.~Yahyaoui and A.~Al-Mutairi, ``A feature-based trust sequence classification
  algorithm,'' \emph{Information Sciences}, vol. 328, pp. 455--484, 2016.

\bibitem{lee2015multi}
C.-H. Lee, ``A multi-phase approach for classifying multi-dimensional sequence
  data,'' \emph{Intelligent Data Analysis}, vol.~19, no.~3, pp. 547--561, 2015.

\bibitem{egho2017user}
E.~Egho, D.~Gay, M.~Boull{\'e}, N.~Voisine, and F.~Cl{\'e}rot, ``A user
  parameter-free approach for mining robust sequential classification rules,''
  \emph{Knowledge and Information Systems}, vol.~52, no.~1, pp. 53--81, 2017.

\bibitem{ntagiou2017protein}
A.~N. Ntagiou, M.~G. Tsipouras, N.~Giannakeas, and A.~T. Tzallas, ``Protein
  structure recognition by means of sequential pattern mining,'' in \emph{2017
  IEEE 17th International Conference on Bioinformatics and Bioengineering
  (BIBE)}.\hskip 1em plus 0.5em minus 0.4em\relax IEEE, 2017, pp. 334--339.

\bibitem{tsai2015pso}
C.-Y. Tsai and C.-J. Chen, ``A pso-ab classifier for solving sequence
  classification problems,'' \emph{Applied Soft Computing}, vol.~27, pp.
  11--27, 2015.

\bibitem{batal2013temporal}
I.~Batal, H.~Valizadegan, G.~F. Cooper, and M.~Hauskrecht, ``A temporal pattern
  mining approach for classifying electronic health record data,'' \emph{ACM
  Transactions on Intelligent Systems and Technology (TIST)}, vol.~4, no.~4,
  p.~63, 2013.

\bibitem{tsai2013time}
C.-Y. Tsai, C.-J. Chen, and C.-J. Chien, ``A time-interval sequence
  classification method,'' \emph{Knowledge and Information Systems}, vol.~37,
  no.~2, pp. 251--278, 2013.

\bibitem{exarchos2009optimized}
T.~P. Exarchos, M.~G. Tsipouras, C.~Papaloukas, and D.~I. Fotiadis, ``An
  optimized sequential pattern matching methodology for sequence
  classification,'' \emph{Knowledge and Information Systems}, vol.~19, no.~2,
  pp. 249--264, 2009.

\bibitem{tseng2005cbs}
V.~S. Tseng and C.-H. Lee, ``Cbs: A new classification method by using
  sequential patterns,'' in \emph{Proceedings of the 2005 SIAM International
  Conference on Data Mining}.\hskip 1em plus 0.5em minus 0.4em\relax SIAM,
  2005, pp. 596--600.

\bibitem{tseng2009effective}
------, ``Effective temporal data classification by integrating sequential
  pattern mining and probabilistic induction,'' \emph{Expert Systems with
  Applications}, vol.~36, no.~5, pp. 9524--9532, 2009.

\bibitem{syed2009learning}
Z.~Syed, P.~Indyk, and J.~Guttag, ``Learning approximate sequential patterns
  for classification,'' \emph{Journal of Machine Learning Research}, vol.~10,
  no.~8, pp. 1913--1936, 2009.

\bibitem{lesh1999mining}
N.~Lesh, M.~J. Zaki, and M.~Ogihara, ``Mining features for sequence
  classification,'' in \emph{Proceedings of the 5th ACM SIGKDD International
  Conference on Knowledge Discovery and Data Mining}.\hskip 1em plus 0.5em
  minus 0.4em\relax ACM, 1999, pp. 342--346.

\bibitem{rani2008rbnbc}
P.~Rani and V.~Pudi, ``Rbnbc: Repeat based na{\"\i}ve bayes classifier for
  biological sequences,'' in \emph{Proceedings of the 8th IEEE International
  Conference on Data Mining}.\hskip 1em plus 0.5em minus 0.4em\relax IEEE,
  2008, pp. 989--994.

\bibitem{holat2014sequence}
P.~Holat, M.~Plantevit, C.~Ra{\"\i}ssi, N.~Tomeh, T.~Charnois, and
  B.~Cr{\'e}milleux, ``Sequence classification based on delta-free sequential
  patterns,'' in \emph{Proceedings of the 2014 IEEE International Conference on
  Data Mining}.\hskip 1em plus 0.5em minus 0.4em\relax IEEE, 2014, pp.
  170--179.

\bibitem{febrer2016spac}
J.~K. Febrer-Hern{\'a}ndez, R.~Hern{\'a}ndez-Le{\'o}n, C.~Feregrino-Uribe, and
  J.~Hern{\'a}ndez-Palancar, ``Spac-nf: A classifier based on sequential
  patterns with high netconf,'' \emph{Intelligent Data Analysis}, vol.~20,
  no.~5, pp. 1101--1113, 2016.

\bibitem{he2018mining}
Z.~He, S.~Zhang, F.~Gu, and J.~Wu, ``Mining conditional discriminative
  sequential patterns,'' \emph{Information Sciences}, vol. 478, pp. 524--539,
  2019.

\bibitem{ifrim2008fast}
G.~Ifrim, G.~Bakir, and G.~Weikum, ``Fast logistic regression for text
  categorization with variable-length n-grams,'' in \emph{Proceedings of the
  14th ACM SIGKDD International Conference on Knowledge Discovery and Data
  Mining}.\hskip 1em plus 0.5em minus 0.4em\relax ACM, 2008, pp. 354--362.

\bibitem{ifrim2011bounded}
G.~Ifrim and C.~Wiuf, ``Bounded coordinate-descent for biological sequence
  classification in high dimensional predictor space,'' in \emph{Proceedings of
  the 17th ACM SIGKDD International Conference on Knowledge Discovery and Data
  Mining}.\hskip 1em plus 0.5em minus 0.4em\relax ACM, 2011, pp. 708--716.

\bibitem{okanohara2009text}
D.~Okanohara and J.~Tsujii, ``Text categorization with all substring
  features,'' in \emph{Proceedings of the 2009 SIAM International Conference on
  Data Mining}.\hskip 1em plus 0.5em minus 0.4em\relax SIAM, 2009, pp.
  838--846.

\bibitem{sonnenburg2005learning}
S.~Sonnenburg, G.~R{\"a}tsch, and C.~Sch{\"a}fer, ``Learning interpretable svms
  for biological sequence classification,'' in \emph{Proceedings of the 9th
  Annual International Conference on Research in Computational Molecular
  Biology}.\hskip 1em plus 0.5em minus 0.4em\relax Berlin, Germany: Springer,
  2005, pp. 389--407.

\bibitem{leslie2001spectrum}
C.~Leslie, E.~Eskin, and W.~S. Noble, ``The spectrum kernel: A string kernel
  for svm protein classification,'' in \emph{Proceedings of the 2002 Paciﬁc
  Symposium on Biocomputing}.\hskip 1em plus 0.5em minus 0.4em\relax World
  Scientific, 2002, pp. 564--575.

\bibitem{lodhi2002text}
H.~Lodhi, C.~Saunders, J.~Shawe-Taylor, N.~Cristianini, and C.~Watkins, ``Text
  classification using string kernels,'' \emph{Journal of Machine Learning
  Research}, vol.~2, no.~2, pp. 419--444, 2002.

\bibitem{eskin2003mismatch}
E.~Eskin, J.~Weston, W.~S. Noble, and C.~S. Leslie, ``Mismatch string kernels
  for svm protein classification,'' in \emph{Advances in Neural Information
  Processing Systems}, 2003, pp. 1441--1448.

\bibitem{leslie2004fast}
C.~Leslie and R.~Kuang, ``Fast string kernels using inexact matching for
  protein sequences,'' \emph{Journal of Machine Learning Research}, vol.~5,
  no.~11, pp. 1435--1455, 2004.

\bibitem{mikolov2013distributed}
T.~Mikolov, I.~Sutskever, K.~Chen, G.~S. Corrado, and J.~Dean, ``Distributed
  representations of words and phrases and their compositionality,'' in
  \emph{Advances in Neural Information Processing Systems}, 2013, pp.
  3111--3119.

\bibitem{le2014distributed}
Q.~Le and T.~Mikolov, ``Distributed representations of sentences and
  documents,'' in \emph{Proceedings of the 2014 International Conference on
  Machine Learning}, 2014, pp. 1188--1196.

\bibitem{nguyen2018sqn2vec}
D.~Nguyen, W.~Luo, T.~D. Nguyen, S.~Venkatesh, and D.~Phung, ``Sqn2vec:
  Learning sequence representation via sequential patterns with a gap
  constraint,'' in \emph{Proceedings of the 2018 Joint European Conference on
  Machine Learning and Knowledge Discovery in Databases}.\hskip 1em plus 0.5em
  minus 0.4em\relax Cham, Switzerland: Springer, 2018, pp. 569--584.

\bibitem{iosifidis2013multidimensional}
A.~Iosifidis, A.~Tefas, and I.~Pitas, ``Multidimensional sequence
  classification based on fuzzy distances and discriminant analysis,''
  \emph{IEEE Transactions on Knowledge and Data Engineering}, vol.~25, no.~11,
  pp. 2564--2575, 2013.

\bibitem{kate2016using}
R.~J. Kate, ``Using dynamic time warping distances as features for improved
  time series classification,'' \emph{Data Mining and Knowledge Discovery},
  vol.~30, no.~2, pp. 283--312, 2016.

\bibitem{blackshields2008fast}
G.~Blackshields, M.~Larkin, I.~M. Wallace, A.~Wilm, and D.~G. Higgins, ``Fast
  embedding methods for clustering tens of thousands of sequences,''
  \emph{Computational Biology and Chemistry}, vol.~32, no.~4, pp. 282--286,
  2008.

\bibitem{voevodski2012active}
K.~Voevodski, M.-F. Balcan, H.~R{\"o}glin, S.-H. Teng, and Y.~Xia, ``Active
  clustering of biological sequences,'' \emph{Journal of Machine Learning
  Research}, vol.~13, no.~1, pp. 203--225, 2012.

\bibitem{Faloutsos1995FastMap}
C.~Faloutsos and K.-I. Lin, ``Fastmap: A fast algorithm for indexing,
  data-mining and visualization of traditional and multimedia datasets,'' in
  \emph{Proceedings of the 1995 ACM SIGMOD International Conference on
  Management of Data}, 1995, pp. 163--174.

\bibitem{hjaltason2003properties}
G.~R. Hjaltason and H.~Samet, ``Properties of embedding methods for similarity
  searching in metric spaces,'' \emph{IEEE Transactions on Pattern Analysis and
  Machine Intelligence}, vol.~25, no.~5, pp. 530--549, 2003.

\bibitem{liu2014discriminative}
X.~Liu, J.~Wu, F.~Gu, J.~Wang, and Z.~He, ``Discriminative pattern mining and
  its applications in bioinformatics,'' \emph{Briefings in Bioinformatics},
  vol.~16, no.~5, pp. 884--900, 2015.

\bibitem{zhang2017up}
C.~Zhang, C.~Liu, X.~Zhang, and G.~Almpanidis, ``An up-to-date comparison of
  state-of-the-art classification algorithms,'' \emph{Expert Systems with
  Applications}, vol.~82, pp. 128--150, 2017.

\bibitem{jain1999data}
A.~K. Jain, M.~N. Murty, and P.~J. Flynn, ``Data clustering: a review,''
  \emph{ACM Computing Surveys}, vol.~31, no.~3, pp. 264--323, 1999.

\bibitem{society2014novel}
T.~X. Society, S.~Wang, Q.~Jiang, and J.~Z. Huang, ``A novel variable-order
  markov model for clustering categorical sequences,'' \emph{IEEE Transactions
  on Knowledge and Data Engineering}, vol.~26, no.~10, pp. 2339--2353, 2014.

\bibitem{willett1988recent}
P.~Willett, ``Recent trends in hierarchic document clustering: a critical
  review,'' \emph{Information Processing and Management}, vol.~24, no.~5, pp.
  577--597, 1988.

\bibitem{rieck2008linear}
K.~Rieck and P.~Laskov, ``Linear-time computation of similarity measures for
  sequential data,'' \emph{Journal of Machine Learning Research}, vol.~9,
  no.~1, pp. 23--48, 2008.

\bibitem{gibbons2011nonparametric}
J.~D. Gibbons and S.~Chakraborti, \emph{Nonparametric statistical
  inference}.\hskip 1em plus 0.5em minus 0.4em\relax Berlin, Germany: Springer,
  2011.

\bibitem{benjamini1995controlling}
Y.~Benjamini and Y.~Hochberg, ``Controlling the false discovery rate: a
  practical and powerful approach to multiple testing,'' \emph{Journal of the
  Royal Statistical Society: Series B (Methodological)}, vol.~57, no.~1, pp.
  289--300, 1995.

\bibitem{mann1947test}
H.~B. Mann and D.~R. Whitney, ``On a test of whether one of two random
  variables is stochastically larger than the other,'' \emph{The Annals of
  Mathematical Statistics}, vol.~18, no.~1, pp. 50--60, 1947.

\bibitem{lichman2013uci}
\BIBentryALTinterwordspacing
M.~Lichman, ``Uci machine learning repository, 2013,'' University of
  California, Irvine, School of Information and Computer Sciences, 2013.
  [Online]. Available: \url{http://archive.ics.uci.edu/ml}
\BIBentrySTDinterwordspacing

\bibitem{mantyjarvi2004sensor}
J.~M{\"a}ntyj{\"a}rvi, J.~Himberg, P.~Kangas, U.~Tuomela, and P.~Huuskonen,
  ``Sensor signal data set for exploring context recognition of mobile
  devices,'' in \emph{Proceedings of the 2nd International Conference on
  Pervasive Computing}, 2004, pp. 18--23.

\bibitem{wei2014improved}
L.~Wei, M.~Liao, Y.~Gao, R.~Ji, Z.~He, and Q.~Zou, ``Improved and promising
  identification of human micrornas by incorporating a high-quality negative
  set,'' \emph{IEEE/ACM Transactions on Computational Biology and
  Bioinformatics}, vol.~11, no.~1, pp. 192--201, 2014.

\bibitem{kim2014convolutional}
Y.~Kim, ``Convolutional neural networks for sentence classification,'' in
  \emph{Proceedings of the 2014 Conference on Empirical Methods in Natural
  Language Processing}.\hskip 1em plus 0.5em minus 0.4em\relax Association for
  Computational Linguistics, 2014, pp. 1746--1751.

\bibitem{pei2001prefixspan}
J.~Pei, J.~Han, B.~Mortazavi-Asl, H.~Pinto, Q.~Chen, U.~Dayal, and M.-C. Hsu,
  ``Prefixspan: Mining sequential patterns efficiently by prefix-projected
  pattern growth,'' in \emph{Proceedings 17th International Conference on Data
  Engineering}.\hskip 1em plus 0.5em minus 0.4em\relax IEEE, 2001, pp.
  215--224.

\bibitem{hall2009weka}
M.~Hall, E.~Frank, G.~Holmes, B.~Pfahringer, P.~Reutemann, and I.~H. Witten,
  ``The weka data mining software: an update,'' \emph{ACM SIGKDD Explorations
  Newsletter}, vol.~11, no.~1, pp. 10--18, 2009.

\bibitem{pedregosa2011scikit}
F.~Pedregosa, G.~Varoquaux, A.~Gramfort, V.~Michel, B.~Thirion, O.~Grisel,
  M.~Blondel, P.~Prettenhofer, R.~Weiss, V.~Dubourg \emph{et~al.},
  ``Scikit-learn: Machine learning in python,'' \emph{Journal of Machine
  Learning Research}, vol.~12, no.~10, pp. 2825--2830, 2011.

\end{thebibliography}

\begin{IEEEbiography}[{\includegraphics[width=1in,height=1.25in,clip,keepaspectratio]{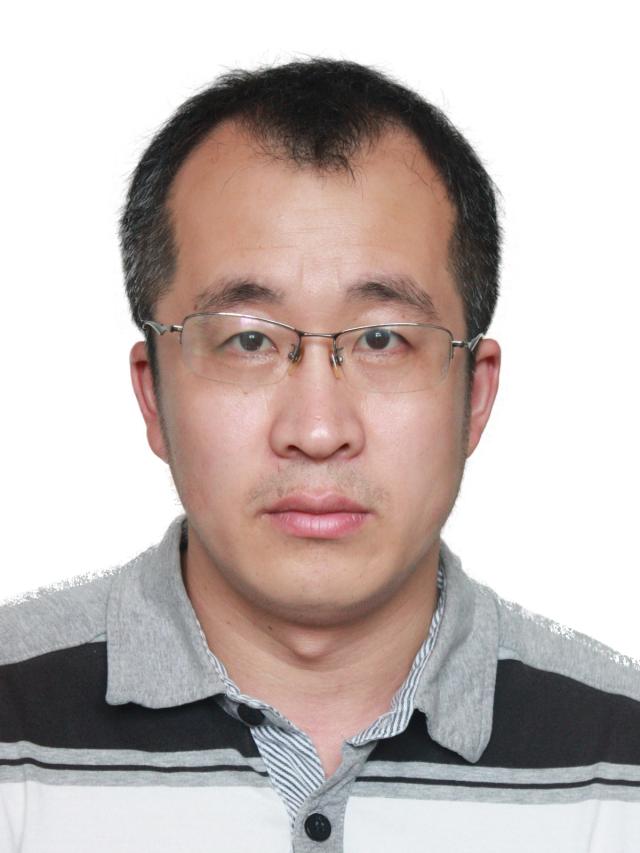}}]{Zengyou He}
received the BS, MS, and PhD
degrees in computer science from Harbin Institute of Technology, China, in 2000, 2002, and
2006, respectively. He was a research associate
in the Department of Electronic and Computer
Engineering, Hong Kong University of Science
and Technology from February 2007 to February
2010. He is currently a professor in the School
of software, Dalian University of Technology. His
research interests include data mining and bioinformatics.
\end{IEEEbiography}

\begin{IEEEbiography}[{\includegraphics[width=1in,height=1.25in,clip,keepaspectratio]{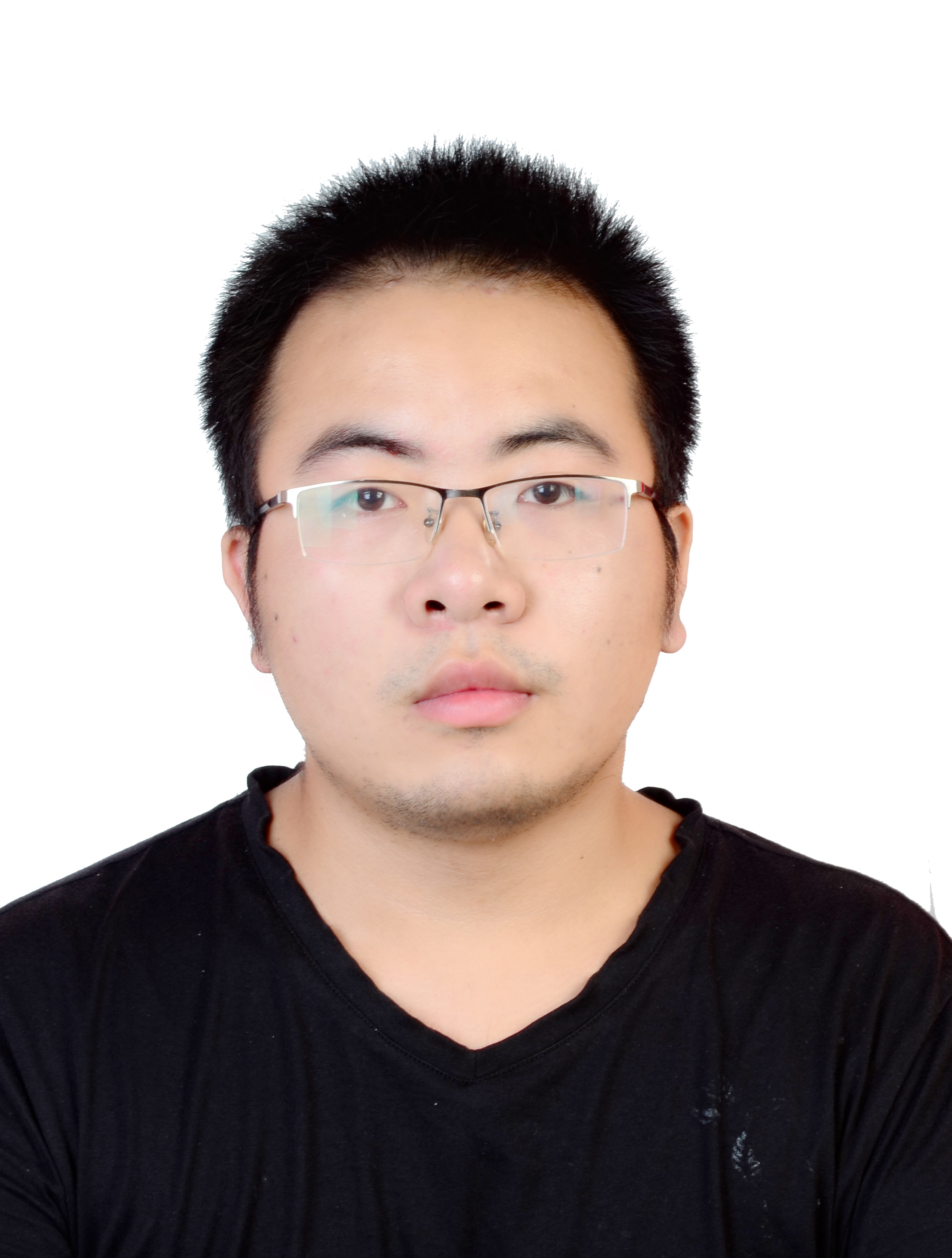}}]{Guangyao Xu}
received the BS degree in electronic and information engineering from Dalian Maritime University, China, in 2018. He is currently working toward the MS degree in the School of Software at Dalian University of Technology. His research interests include data mining and its applications.
\end{IEEEbiography}

\begin{IEEEbiography}[{\includegraphics[width=1in,height=1.25in,clip,keepaspectratio]{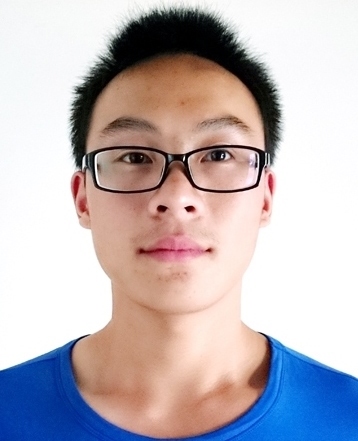}}]{Chaohua Sheng}
received the BS degree in software engineering from Dalian University of Technology, China, in 2018. He is currently working toward the MS degree in the School of Software at Dalian University of Technology. His research interests include data mining and its applications.
\end{IEEEbiography}

\begin{IEEEbiography}[{\includegraphics[width=1in,height=1.25in,clip,keepaspectratio]{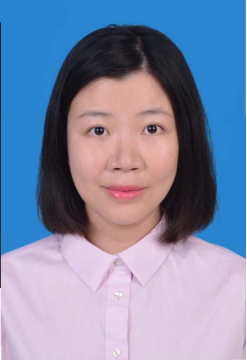}}]{Bo Xu}
received the BSc and PhD degrees from
the Dalian University of Technology, China, in
2007 and 2014, respectively. She is currently
an associate professor in School of Software at the Dalian
University of Technology. Her current research
interests include biomedical literature data mining, information retrieval, and natural language
processing.
\end{IEEEbiography}

\begin{IEEEbiography}[{\includegraphics[width=1in,height=1.25in,clip,keepaspectratio]{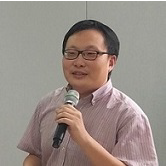}}]{Quan Zou}
Quan Zou (M'13-SM'17) received the
BSc, MSc and the PhD degrees in computer science from Harbin Institute of Technology, China,
in 2004, 2007 and 2009, respectively. He worked
in Xiamen University and Tianjin University from
2009 to 2018 as an assistant professor, associate professor and professor. He is currently a
professor in the Institute of Fundamental and
Frontier Sciences, University of Electronic Science and Technology of China. His research is
in the areas of bioinformatics, machine learning
and parallel computing. Several related works have been published by
Science, Briefings in Bioinformatics, Bioinformatics, IEEE/ACM Transactions on Computational Biology and Bioinformatics, etc. Google scholar
showed that his more than 100 papers have been cited more than 5000
times. He is the editor-in-chief of Current Bioinformatics, associate editor
of IEEE Access, and the editor board member of Computers in Biology
and Medicine, Genes, Scientific Reports, etc. He was selected as one
of the Clarivate Analytics Highly Cited Researchers in 2018.
\end{IEEEbiography}

	\EOD
	
\end{document}